%% file: main.tex
\documentclass[10pt,twocolumn,letterpaper]{article}

\usepackage{cvpr}              

\input{preamble}

\definecolor{cvprblue}{rgb}{0.21,0.49,0.74}
\usepackage[pagebackref,breaklinks,colorlinks,allcolors=cvprblue]{hyperref}

\usepackage[accsupp]{axessibility}  

\usepackage{amsmath}
\usepackage{comment}
\usepackage{xcolor,colortbl}

\definecolor{lightred}{rgb}{1,0.8,0.8}
\definecolor{lightgreen}{rgb}{0.8,1,0.8}
\definecolor{lightblue}{rgb}{0.88,0.96,1}
\definecolor{lightgray}{rgb}{0.9,0.9,0.9}
\newcommand{\bgb}{\cellcolor{lightblue}}

\def\paperID{14519} 
\def\confName{CVPR}
\def\confYear{2026}

\title{\texttt{SPDMark}: Selective Parameter Displacement for Robust Video Watermarking}

\author{Samar Fares$^{1}$ \quad Nurbek Tastan$^{1}$ \quad Karthik Nandakumar$^{1,2}$
\and
$^{1}$Mohamed bin Zayed University of Artificial Intelligence (MBZUAI), Abu Dhabi, UAE\\
$^{2}$Michigan State University (MSU), East Lansing, MI, USA\\
\texttt{\small samar.fares@mbzuai.ac.ae, nurbek.tastan@mbzuai.ac.ae, nandakum@msu.edu}
}

\begin{document}
\maketitle
\input{sec/0_abstract}
\input{sec/1_introduction}
\input{sec/2_background}

\input{sec/3_method}

\input{sec/4_experiments}

\input{sec/5_conclusion}

{
    \small
    \bibliographystyle{ieeenat_fullname}
    \bibliography{main}
}


\input{sec/X_suppl}

\end{document}

%% file: preamble.tex
\usepackage{bbm}
\usepackage{amsmath, amssymb}
\usepackage{multirow}
\usepackage{float}
\usepackage{graphicx,booktabs}
\usepackage{multicol}



%% file: sec/0_abstract.tex
\begin{abstract}
The advent of high-quality video generation models has amplified the need for robust watermarking schemes that can be used to reliably detect and track the provenance of generated videos. Existing video watermarking methods based on both post-hoc and in-generation approaches fail to simultaneously achieve imperceptibility, robustness, and computational efficiency. This work introduces a novel framework for in-generation video watermarking called \textbf{\texttt{SPDMark}} (pronounced `SpeedMark') based on \textbf{selective parameter displacement} of a video diffusion model. Watermarks are embedded into the generated videos by modifying a subset of parameters in the generative model. To make the problem tractable, the displacement is modeled as an additive composition of layer-wise basis shifts, where the final composition is indexed by the watermarking key. For parameter efficiency, this work specifically leverages low-rank adaptation (LoRA) to implement the basis shifts. During the training phase, the basis shifts and the watermark extractor are jointly learned by minimizing a combination of message recovery, perceptual similarity, and temporal consistency losses. To detect and localize temporal modifications in the watermarked videos, we use a cryptographic hashing function to derive frame-specific watermark messages from the given base watermarking key. During watermark extraction, maximum bipartite matching is applied to recover the correct frame order, even from temporally tampered videos. Evaluations on both text-to-video and image-to-video generation models demonstrate the ability of \textbf{\texttt{SPDMark}} to generate imperceptible watermarks that can be recovered with high accuracy and also establish its robustness against a variety of common video modifications. 
Code is \href{https://github.com/Samar-Fares/SPDMark}{here}.

\end{abstract}

%% file: sec/1_introduction.tex
\section{Introduction}

The ability of video generation models to generate realistic and temporally coherent videos has improved dramatically within the past 3 years ~\cite{Stablevideodiffusion,wang2023modelscope,openai2024sora,bao2024vidu}. This raises serious concerns about the provenance of AI-generated videos and the responsible deployment of such generative models. Watermarking has been touted as a practical solution for tracking the provenance of AI-generated content. Recent government regulations such as the EU AI Act \cite{rijsbosch2025adoption} and the U.S. Executive Order on AI \cite{biden2023executive} have also recommended watermarking of AI-generated content to mitigate misuse of such content. An ideal video watermarking must be (i) imperceptible across space and time, (ii) reliably recoverable even after common modifications, and (iii) computationally efficient. Unlike images, videos introduce the added difficulty of retaining \emph{temporal structure}: frame drops, swaps, or insertions can break watermark synchronization even when the spatial quality is preserved.

Existing video watermarking approaches fall into two main groups. \textit{Post-hoc} methods~\cite{videoSeal} operate on the generated videos but add latency and cannot leverage generative priors. In contrast, in-generation techniques embed the watermark during the video generation process. These schemes can be further categorized into \textit{noise-space}~\cite{VideoShield} and \textit{model fine-tuning}~~\cite{VideoMark,LVMark} methods. Noise-space methods embed the watermarking message within the diffusion noise and decode via DDIM inversion~\cite{song2020denoising}, achieving large message capacity at the cost of high computation. \emph{Model fine-tuning} methods partially fine-tune the generative model (typically the latent decoder) to embed the watermark message. However, these methods often apply a uniform modulation or embed a single fixed signature, thereby limiting the detection of temporal manipulations. Existing video watermarking schemes still face trade-offs between imperceptibility, robustness, and efficiency.

This work introduces \textbf{\texttt{SPDMark}}, a scalable in-generation video watermarking scheme based on the concept of \emph{Selective Parameter Displacement} (SPD). Instead of perturbing pixels or noise, \texttt{SPDMark} embeds watermark messages by selectively displacing parameters of the generative model. The displacement of parameters within the frozen generative model is achieved by activating a sparse combination of learned low-rank basis shifts. A single trained dictionary of low-rank basis shifts supports \emph{arbitrary watermark keys} without retraining: each key (and each frame) simply selects a different combination of basis shifts. This yields strong imperceptibility and per-frame watermarking at negligible inference overhead. The main contributions of this work can be summarized as follows:
\begin{itemize}[leftmargin=*,noitemsep]
    \item We propose the Selective Parameter Displacement (\texttt{SPDMark}) framework for enabling multi-key in-generation watermarking in video diffusion models.
    
    \item To practically realize the \texttt{SPDMark} framework, we propose modeling the displacement as an additive composition of layer-wise low-rank basis shifts and determining this composition based on the watermarking key. During training, the dictionary of basis shifts and the watermark extractor are jointly learned by minimizing a combination of message recovery and imperceptibility losses.

    \item We also propose a mechanism for embedding unique watermark messages in every frame of the generated video and a watermark detection procedure based on maximum bipartite matching and statistical hypothesis testing, which allows for localization of temporal modifications applied to the watermarked video. 
    
    \item We demonstrate the robustness of \texttt{SPDMark} and its ability to preserve video  quality through experiments on two video diffusion models and several attack scenarios.
\end{itemize}

%% file: sec/2_background.tex
\section{Background}

\subsection{Preliminaries}

\textbf{Notation}: Let $\mathcal{G}_{\Phi}: \mathcal{Z} \times \mathcal{C} \rightarrow \mathcal{X}$ 
denote a video generation model parameterized by $\Phi$. In general, we assume that the generative model $\mathcal{G}_{\Phi}$  maps latent noise $\mathbf{z} \in \mathcal{Z}$ and conditioning input $\mathbf{c} \in \mathcal{C}$ (e.g., a text prompt or an image) to a video output $\mathbf{x} \in \mathcal{X}$, i.e., $\mathbf{x} = \mathcal{G}_{\Phi}(\mathbf{z},\mathbf{c})$ . A video $\mathbf{x} \in \mathcal{X}$ consists of a sequence of $T$ frames $[x_1,x_2,\cdots,x_T]$, where $x_t$ represents the $t$-th frame in the video ($t \in [1,T]$). A watermarking system $\mathcal{W} = (\mathcal{U}_{\zeta}, \mathcal{V}_{\eta})$ consists of an encoder-extractor pair. The  \textit{encoder} $\mathcal{U}_{\zeta}$ embeds an $M$-bit message $\kappa \in \mathcal{K} \subseteq \{0,1\}^M$ into the generated video, resulting in a watermarked video $\tilde{\mathbf{x}} = \mathcal{U}_{\zeta}(\kappa) || \mathcal{G}_{\Phi}(\mathbf{z},\mathbf{c})$, where $||$ denotes a generic combination of two functions. The watermark \textit{extractor} $\mathcal{V}_{\eta}: \mathcal{X} \rightarrow \mathcal{K}$ attempts to recover the embedded message $\hat{\kappa} = \mathcal{V}_{\eta}(\tilde{\mathbf{x}})$ from a watermarked video $\tilde{\mathbf{x}}$. 

\noindent \textbf{Problem Statement for Video Watermarking}: We consider four participants: (1) The \textit{User} wishes to generate a video conditioned on some input $\mathbf{c}$, where $\mathbf{c}$ is either a text prompt or an image that controls the semantic content of the generated video. (2) The \textit{Model Owner} has $\mathcal{G}_{\Phi}$ and $\mathcal{W}$ and generates a watermarked video $\tilde{\mathbf{x}}$ based on the user's input $\mathbf{c}$. (3) The  \textit{Verifier} uses $\mathcal{V}_{\eta}$ to determine whether the given video contains a valid watermark. (4) The \textit{Adversary} attempts to apply a transformation $\mathcal{A}$ to a video, either to remove the watermark from a valid watermarked video $\tilde{\mathbf{x}}$ or to forge a valid watermark into a non-watermarked video $\mathbf{x}$ (which could be natural or synthetically generated). We assume that the adversary has no access to $\Phi$, $\zeta$, or $\eta$, and that $\mathcal{A}$ is bounded (mostly preserving the visual content). 

An ideal watermarking scheme should satisfy the following four requirements: (i) \textbf{Imperceptibility}: $\tilde{\mathbf{x}}$ should be perceptually indistinguishable from $\mathbf{x}$ (for any $\mathbf{c}$) under visual inspection by the user. (ii) \textbf{Message Recoverability}: Message $\hat{\kappa}$ extracted from valid watermarked videos should match the embedded message $\kappa$ with high probability, and the false positive rate (probability of recovering $\kappa$ from non-watermarked videos) should be low. (iii) \textbf{Robustness}: The watermark should be resistant against small modifications $\mathcal{A}$ (e.g., compression, cropping, frame drop, etc.) applied by the adversary. (iv) \textbf{Computational Efficiency}: The computational effort required to embed and extract the watermark must be small compared to video generation (ideally, offline training effort should also be minimal).
\vspace{-.5em}
\subsection{Image and Video Diffusion Models}
\label{sec:diffusionModels}
Diffusion models~\cite{ho2020denoising,dhariwal2021diffusion} synthesize data by learning to reverse a gradual noising process. Latent Diffusion Models (LDMs)~\cite{rombach2022high} for images improve efficiency by operating in a compressed latent space: an encoder $\mathcal{E}$ maps an image to a latent code, a denoiser (UNet~\cite{ronneberger2015u} or Diffusion Transformer~\cite{peebles2023scalableDit}) iteratively refines the noised latent code, and a decoder $\mathcal{D}$ reconstructs the output image. The encoder-decoder pair can be considered as a Variational Autoencoder (VAE). Video diffusion models extend this pipeline to capture spatiotemporal structure. Early systems such as ModelScope~\cite{wang2023modelscope} and Stable Video Diffusion~\cite{Stablevideodiffusion} pair 2D VAEs with 3D UNets, while recent models including Sora~\cite{openai2024sora} and Vidu~\cite{bao2024vidu} adopt 3D VAEs and DiT backbones to model long-range motion. These architectural differences impact our watermarking system design: schemes that target the diffusion process or the denoiser may transfer less naturally across UNet- and DiT-based models and the set of parameters selected for displacement (encoder, denoiser, or decoder) affects the robustness and verification cost. \texttt{SPDMark} therefore embeds watermarks in the decoder $\mathcal{D}$, a component shared across latent video diffusion models with different denoiser backbones, including both UNet- and Transformer-based variants.

\subsection{Watermarking for Image Generation Models}
Watermarking for generative models generally follows two paradigms: \emph{post-hoc} and \emph{in-generation}. Post-hoc methods~\cite{zhu2018hiddenhidingdatadeep} embed signals after generation using encoder-decoder networks, achieving good imperceptibility but adding latency and remaining decoupled from the generative model. In-generation methods embed watermarks directly in the diffusion process. Noise-space techniques such as Tree-Ring~\cite{wen2023tree} and Gaussian Shading~\cite{yang2024gaussian} modify the initial noise and decode via DDIM inversion~\cite{song2020denoising}; however, inversion is computationally expensive and fragile under perturbations. Model fine-tuning approaches avoid inversion by modifying model weights: Stable Signature~\cite{fernandez2023stable} and WOUAF~\cite{kim2024wouaf} fine-tune diffusion decoders, and AQuaLoRA~\cite{feng2024aqualora} uses LoRA modules for image-level embedding. While these methods offer strong imperceptibility for images, they do not address the temporal coherence required in video watermarking.

\subsection{Video Watermarking}
Since video watermarking requires preserving both spatial imperceptibility and temporal coherence, extensions of image watermarking approaches have been proposed for videos. \textbf{Noise-space methods}: VideoShield~\cite{VideoShield} perturbs the initial diffusion noise and decodes messages via DDIM inversion, enabling tamper localization but incurring high computational cost and remaining vulnerable to temporal manipulations such as frame reordering.
VideoMark~\cite{VideoMark} adds per-frame pseudo-random noise with error-correcting codes to improve matching, but still inherits the fragility and overhead of full inversion. \textbf{Model fine-tuning methods}: LVMark~\cite{LVMark} jointly trains a modified latent decoder and 3D extractor, achieving strong robustness but modulating all layers uniformly, which limits per-frame control and harms visual quality. VideoSignature (VidSig)~\cite{VideoSignature} freezes perceptually sensitive layers (PAS) and adds a temporal alignment module, but embeds a single fixed signature rather than frame-specific messages. \textbf{Post-hoc methods}: VideoSeal~\cite{videoSeal} embeds watermarks after generation using a lightweight U-Net and ViT extractor, aided by differentiable augmentations and codec simulation. While effective for compression, its post-generation nature adds latency and cannot leverage generative priors. \texttt{SPDMark} differs from the above methods by learning a fixed dictionary of low-rank basis shifts that can be \emph{dynamically composed per frame} according to arbitrary binary messages, enabling efficient multi-key watermarking with per-frame control and without per-key retraining. \texttt{SPDMark} is also computationally efficient because it avoids inversion entirely.

%% file: sec/3_method.tex
\section{Proposed Method: \texttt{SPDMark}}
\label{sec:method}

\begin{figure*}[t]
    \centering
    \includegraphics[width=\linewidth]{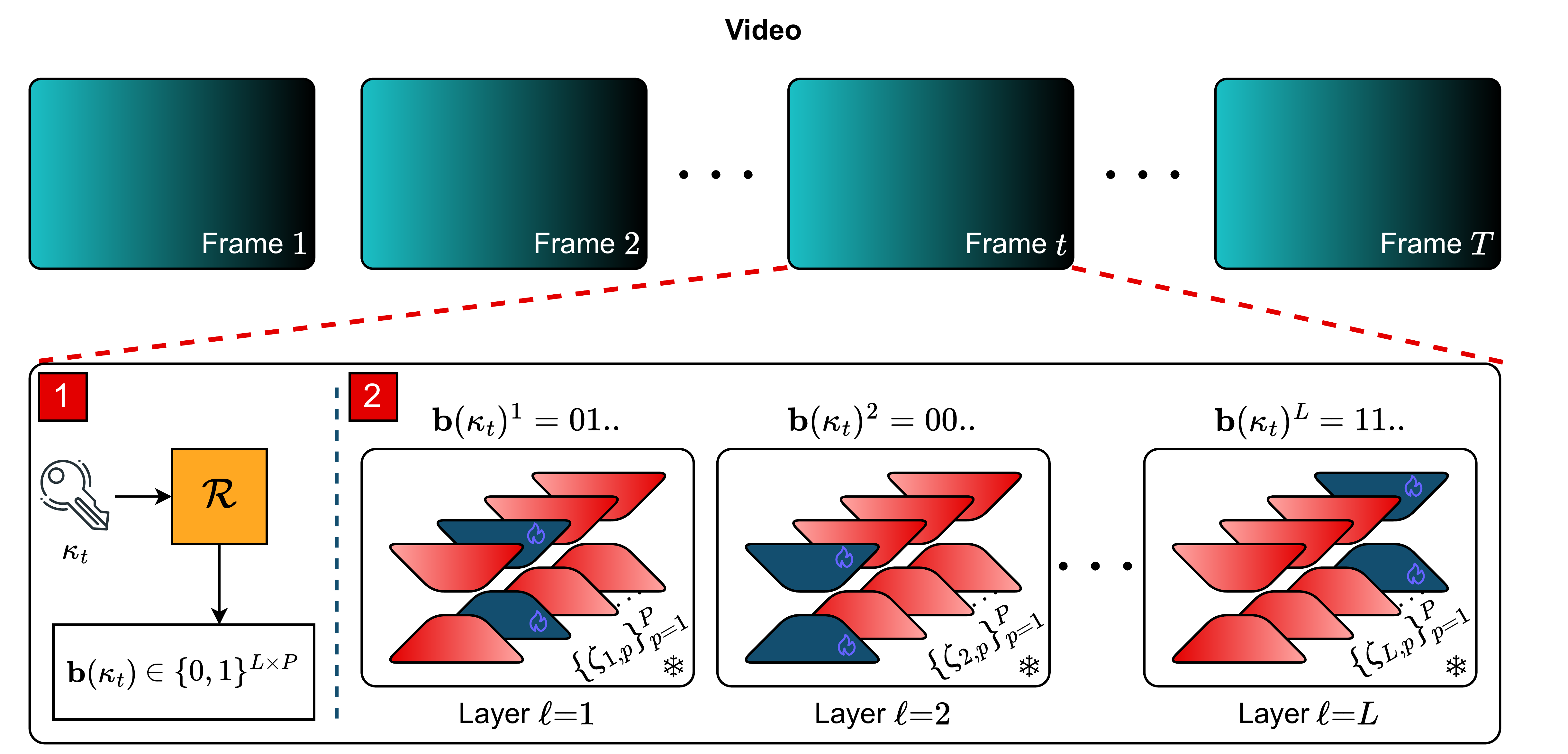}
\caption{\textbf{\texttt{SPDMark} pipeline.}
    Watermarking key $\kappa$ is expanded into per-frame messages $\{\kappa_t\}$.
    Each $\kappa_t$ is mapped to a binary mask $\mathbf{b}(\kappa_t)$, yielding the parameter
    displacement $\Delta\Phi_M(\kappa_t)=\mathbf{b}(\kappa_t)\!\otimes\!\zeta$, which is applied to the frozen decoder to produce the watermarked video $\tilde{\mathbf{x}}$. The watermark extractor $\mathcal{V}_\eta$ recovers the messages frame-wise; verification uses graph-based alignment and hypothesis testing (Sec.~\ref{sec:SPDimplemetation}).}
    \label{fig:spdmark-framework}
\end{figure*}

\texttt{SPDMark} is an in-generation video watermarking method, where the core idea is to embed the watermark by selectively modifying the parameters of the given video diffusion model and jointly learn a lightweight watermark extractor.  

\subsection{Selective Parameter Displacement Framework}

Let $\mathcal{G}_{\Phi}$ be the given video generation model and $\kappa$ be the given watermarking key. Let $\mathbf{c}$ be any conditioning input provided by the user and $\mathbf{z}$ be any latent noise chosen by the model owner. The objective of the parameter displacement framework is to displace the parameters $\Phi$ of $\mathcal{G}$ such that the videos generated by the displaced model embed the watermark key $\kappa$. Let $(\Phi + \Delta \Phi)$ be the parameters of the displaced model, where $\Delta \Phi$ denotes the displacement. Since $\Delta \Phi$ encodes the key $\kappa$, the mapping from $\kappa$ to $\Delta \Phi$ can be considered as the watermarker encoder. The challenge is to learn this encoder carefully so that the applied displacement does not affect the visual quality of the watermarked videos. Therefore, we need to design the watermarking system $\mathcal{W}$ and learn the displacement such that:
\begin{equation}
\label{eq:mainopt}
    \min_{\zeta,\eta} \mathcal{L}_{\textrm{imp}}(\mathbf{x},\tilde{\mathbf{x}}) + \mathcal{L}_{\textrm{rec}}(\mathcal{V}_\eta(\tilde{\mathbf{x}}),\kappa),
\end{equation}
\noindent where $\tilde{\mathbf{x}} = \mathcal{G}_{\Phi+\Delta \Phi}(\mathbf{z},\mathbf{c})$ and $\Delta \Phi = \mathcal{U}_{\zeta}(\kappa)$. In the above formulation, $\mathcal{L}_{\textrm{imp}}$ and $\mathcal{L}_{\textrm{rec}}$ are loss functions that enforce the imperceptibility and message recoverability requirements, respectively.

To make the problem of learning the parameter displacement $\Delta \Phi$ tractable, we first partition the parameters of the given generative model into two components $\Phi_U$ and $\Phi_M$, where $\Phi_U$ denotes the parameters that remain untouched and $\Phi_M$ denotes the parameters to be modified. Correspondingly, the displacement $\Delta \Phi$ is also split into two components, where the first component is zero and the second component  $\Delta \Phi_M$ needs to be learned.
\begin{equation}
    \Phi = \begin{bmatrix} \Phi_U \\ \Phi_M\end{bmatrix}, \Delta \Phi = \begin{bmatrix}
        \mathbf{0} \\ \Delta \Phi_M
    \end{bmatrix}.
\end{equation}
Next, we assume that the parameters to be modified ($\Phi_M$) are spread across $L$ distinct layers in the generative model. Let $\phi_{\ell}$ denote the parameters of layer $\ell$, $\ell \in [1,L]$ and $\Delta \phi_{\ell}$ denote the corresponding displacement. Hence,
\begin{equation}
    \Phi_M = \begin{bmatrix} \phi_1 \\ \phi_2 \\ \vdots \\ \phi_L \end{bmatrix}, \Delta \Phi_M = \begin{bmatrix}
        \Delta \phi_1 \\ \Delta \phi_2 \\ \vdots \\ \Delta \phi_L 
    \end{bmatrix}.
\end{equation}
In each layer $\ell \in [1,L]$, the displacement $\Delta \phi_{\ell}$ is further modeled as an \textbf{additive composition of} $P$ \textbf{\textit{basis shifts}}. Let $\{\zeta_{\ell,p}\}_{p=1}^P$ denote the set of basis shifts in layer $\ell$. Therefore,

\begin{equation}
    \Delta \phi_{\ell} = \sum_{p=1}^{P} b_{\ell,p} \zeta_{\ell,p},
\end{equation}

\noindent where $b_{\ell,p} \in \{0,1\}$ is a binary mask indicating if the corresponding basis shift is selected. Let $\zeta = \{\{\zeta_{\ell,p}\}_{p=1}^P\}_{\ell=1}^{L}$ denote the set of all basis shifts across all layers and $\mathbf{b} =  \{\{b_{\ell,p}\}_{p=1}^P\}_{\ell=1}^{L} \in \{0,1\}^{L \times P}$ be the set of all mask values. Since our goal is to define the displacement $\Delta \Phi$ as a function of the watermark key $\kappa$, we employ a key mapping procedure to determine the binary mask using $\kappa$. Let $\mathbf{b}(\kappa)$ be the binary mask indexed by $\kappa$. Thus,
\begin{equation}
    \Delta \Phi_{M} = \mathbf{b}(\kappa) \otimes \zeta = \begin{bmatrix} \sum_{p=1}^{P} b_{1,p} \zeta_{1,p} \\ \sum_{p=1}^{P} b_{2,p} \zeta_{2,p} \\ \vdots \\ \sum_{p=1}^{P} b_{L,p} \zeta_{L,p} \end{bmatrix}.
\end{equation}
\noindent Equations (1) through (5) together define the proposed Selective Parameter Displacement framework for video watermarking (\texttt{SPDMark}).

\subsection{Practical Implementation of \textbf{\texttt{SPDMark}}}
\label{sec:SPDimplemetation}

To realize the \texttt{SPDMark} framework in practice, we make the following implementation choices.

\noindent \textbf{Key Mapping}: Though there are many ways to map the watermark key $\kappa$ to the binary mask $\mathbf{b}$, we use the following simple procedure. Initially, we set all the binary masks $b_{\ell,p}$ to zero ($\forall~ \ell \in [1,L], p \in [1, P]$). We assume that the key length is $M = L\log_2{P}$ bits. Then, we partition the key $\kappa$ into $L$ chunks of $\log_2{P}$ bits, i.e., $\kappa = [\kappa_1, \kappa_2, \cdots, \kappa_L]$, where each  $\kappa_{\ell}$ contains $\log_2{P}$ bits. Let $i_{\ell} = bin2dec(\kappa_{\ell})$ be the decimal representation of $\kappa_{\ell}$. Note that $i_{\ell} \in [0,P-1]$. We set $b_{\ell,i_{\ell}+1}$ to value $1$ in each layer $\ell$. Thus, the key $\kappa$ gets mapped to the binary mask $\mathbf{b}$ and consequently, the displacement $\Delta \Phi$ becomes a function of $\kappa$.

\begin{equation}
    \Delta \Phi_{M}(\kappa) = \mathbf{b}(\kappa) \otimes \zeta = \begin{bmatrix} \zeta_{1,i_{1}+1} \\ \zeta_{2,i_{2}+1} \\ \vdots \\ \zeta_{L,i_{L}+1} \end{bmatrix},
\end{equation}

\noindent where $i_{\ell} = bin2dec(\kappa_{\ell})$. Note that the above key mapping procedure selects only a single basis shift in each layer $\ell$ for parameter displacement.

\noindent \textbf{Layer Selection}: This work assumes that the video generation model is a latent diffusion model, consisting of an encoder, a denoiser, and a decoder, as explained in Section \ref{sec:diffusionModels}. To preserve the quality of video generation, we leave the parameters of the encoder and denoiser untouched and apply SPD only to the diffusion model decoder. 

\noindent \textbf{Choice of basis shifts}: While it is possible to learn the basis shifts $\zeta$ directly, such an approach is not parameter efficient. Hence, we borrow the idea of low-rank adaptation (LoRA) \cite{hu2022lora} and instantiate the basis shifts using low-rank matrix decompositions. Each basis shift $\zeta_{\ell,p}$ is modeled as a low-rank update to the layer parameters as follows.
\begin{equation}
\label{eq:lora-basis}
\begin{aligned}
\zeta_{\ell,p} &= A_{\ell,p}B_{\ell,p},\\[-2pt]
A_{\ell,p} & \in \mathbb{R}^{d\times r},\quad
B_{\ell,p} \in \mathbb{R}^{r\times d},\quad
r \ll d.
\end{aligned}
\end{equation}
\noindent Let $\mathbf{h}_{\ell-1}$ be the input to layer $\ell$ in the original diffusion model and let $\mathbf{h}_{\ell} = \mathcal{F}_{\phi_\ell}(\mathbf{h}_{\ell-1})$ be its output. After parameter displacement, the output of layer $\ell$ is computed as:
\begin{equation}
     \mathbf{h}_{\ell} = \mathcal{F}_{\phi_\ell}(\mathbf{h}_{\ell-1}) + \alpha \mathcal{F}_{\Delta \phi_\ell}(\mathbf{h}_{\ell-1}),
\end{equation}
\noindent where $\Delta \phi_\ell = \zeta_{\ell,p^*} = A_{\ell,p^*}B_{\ell,p^*}$, $p^* = (i_{\ell}+1)$ denotes the basis shift for layer $\ell$ selected based on $\kappa$, and $\alpha$ is a fixed scalar.

\noindent \textbf{Per-Frame Watermark Message Generation}: To detect and localize temporal modifications in the watermarked video, it is essential to embed unique watermark messages in each frame of the video. For a $T$-frame video, we derive frame-specific watermark messages from a video-level secret key $K_{\text{base}}$ using a cryptographic hash function $\mathcal{H}$ (e.g., HMAC-SHA256):
\begin{equation}
\label{eq:perframemsg}
  \kappa_t=\mathrm{Trunc}_M\!\big(\mathcal{H}(K_{\text{base}}, t)\big), \qquad t=1,\ldots,T.
\end{equation}
\noindent where $\mathrm{Trunc}_M$ denotes that the resulting hash value is truncated to the first $M$ bits. Consequently, the parameter displacement is also frame-specific.

\noindent \textbf{Watermark Extractor}: Since the embedded watermark message is frame-specific, the watermark extractor $\mathcal{V}_{\eta}$ is designed to operate on individual frames of the video. Specifically, we employ a ResNet-50~\cite{deepResidual} model pretrained on ImageNet~\cite{deng2009imagenet} and replace the final fully connected layer with a linear head that outputs $M$ logits per frame. 

\noindent \textbf{Loss Functions}: To enforce the message recoverability requirement, we employ the binary cross entropy with logits loss ($\mathrm{BCE}_{\text{logits}}$). Specifically, for a watermarked video $\tilde{\textbf{x}}$,
\begin{equation}
\label{eq:lrec}
\mathcal{L}_{\textrm{rec}}(\mathcal{V}_\eta(\tilde{\mathbf{x}}),\kappa)
\;=\;
\mathbb{E}_{t \sim T}\!\left[
  \mathrm{BCE}_{\text{logits}}\big(\mathcal{V}_{\eta}(\tilde{x}_t),\,\kappa_t\big)
\right],
\end{equation}

\noindent where $\tilde{x}_t$ is the $t$-th frame in $\tilde{\textbf{x}}$ and $\kappa_t$ is the frame-specific watermark message.

To enforce the imperceptibility constraint, we use a weighted combination of perceptual similarity (defined based on LPIPS~\cite{zhang2018perceptual}) and temporal consistency losses. While the perceptual similarity (PS) loss encourages the watermarked video to stay closer to the corresponding original video without watermark, the temporal consistency (TC) loss ensures smooth transition between successive frames of the video (thereby avoiding the flickering effect). Let $\mathbf{x}=\mathcal{G}_{\Phi}(\mathbf{z},\mathbf{c})$ be the non-watermarked video generated by the original model and $\tilde{\mathbf{x}}=\mathcal{G}_{\Phi+\Delta \Phi}(\mathbf{z},\mathbf{c})$ be the watermarked video generated by the displaced model with the same $\mathbf{z}$ and $\mathbf{c}$. The imperceptibility loss is defined as:
\begin{equation}
\label{eq:limp}
\begin{aligned}
    \mathcal{L}_{\textrm{imp}}(\mathbf{x},\tilde{\mathbf{x}})
    &=
    \lambda_{ps} \, \mathbb{E}_{t \sim T}\!\left[
      \mathrm{LPIPS}\big(x_t,\tilde{x}_t\big)\right] \\
    &\quad + \lambda_{tc} \, \mathbb{E}_{t \sim T}\!\left[
      \big\|\delta y_t - \delta\tilde{y}_t\big\|_{1}\right].
\end{aligned}
\end{equation}
\noindent where $y = (0.299\,x^{(R)}+0.587\,x^{(G)}+0.114\,x^{(B)})$ represents the luminance of an image whose RGB channels are denoted as $x^{(R)}$, $x^{(G)}$, and $x^{(B)}$, respectively. Furthermore, $\delta y_t = (y_t - y_{t-1})$ and $\delta\tilde{y}_t = (\tilde{y}_t - \tilde{y}_{t-1})$ denote the luminance differences between successive frames in the non-watermarked and watermarked videos, respectively. Here, $\lambda_{ps}$ and $\lambda_{tc}$ are the weights assigned to the PS and TC losses, respectively. For the TC loss, the expectation is computed over $(T-1)$ frame differences. Since the watermarking procedure must be agnostic to the watermark key $\kappa$, input condition $\mathbf{c}$, and latent noise $\mathbf{z}$, the optimization in Eq. \ref{eq:mainopt} (based on losses defined in Eqs. \ref{eq:lrec} and \ref{eq:limp}) is performed in expectation over $\kappa \sim \mathcal{K}$, $\mathbf{c} \sim \mathcal{C}$, and $\mathbf{z} \sim \mathcal{Z}$. 

\noindent \textbf{Verification of Watermark Validity}: During verification, the verifier applies the watermark extractor $\mathcal{V}_{\eta}$ to the individual frames in the given video $\mathbf{x}^{*}$ to recover a sequence of watermark messages $\mathbf{\hat{K}} = [\hat{\kappa}_1, \hat{\kappa}_2, \cdots, \hat{\kappa}_{T_r}]$, where $T_r$ is the number of frames in the given video. To verify whether $\mathbf{x}^{*}$ contains a valid watermark, the verifier needs access to the video base key $K_{\text{base}}$ used by the model owner for generating the watermarked video. Given $K_{\text{base}}$ and the number of frames $T$ in the original watermarked video, it is straightforward to regenerate the frame-specific messages $\mathbf{K} = [\kappa_1, \kappa_2, \cdots, \kappa_T]$ used by the model owner using Eq.~\ref{eq:perframemsg}. Note that due to temporal modifications (e.g., frame drops, insertions, reordering, etc.) $T_r$ may not be equal to $T$ and there may be temporal misalignment between the frames. We pose the problem of finding matching frames as a \textit{maximum bipartite graph matching} problem and solve it using the Hungarian algorithm \cite{Kuhn1955}. Specifically, the set of messages in $\mathbf{K}$ and  $\mathbf{\hat{K}}$ are modeled as a bipartite graph with the edge weights between vertices in $\mathbf{K}$ and  $\mathbf{\hat{K}}$ defined as:
\begin{equation}
    \bar{S}_{m,n} = 1 - \frac{\psi(\kappa_m,\hat{\kappa}_n)}{M},
\end{equation}
\noindent where $\psi$ is the Hamming distance between two binary strings of length $M$, $m \in [1,T]$, and $n \in [1,T_r]$. Note that the edge weights $\bar{S}_{m,n}$ represent the similarity (proportion of matched bits) between messages $\kappa_m$  and $\hat{\kappa}_n$. We then compute a one-to-one alignment between the reference messages and the extracted messages using maximum-weight bipartite matching.
\begin{equation}
(\pi^*, \rho^*) = \arg\max_{\pi, \rho} \sum_{i} \bar{S}_{\pi_i, \rho_i}.
\vspace{-1em}
\end{equation}
\noindent Let $\mathcal{M} = \{(\pi_i, \rho_i)\}_{i=1}^{|\mathcal{M}|}$ denote the set of assignments made by the Hungarian algorithm, where $|\mathcal{M}|$ is the number of assignments. To verify the validity of each frame assignment, we perform the following hypothesis test. Let $S_{m,n} = (M - \psi(\kappa_m,\hat{\kappa}_n))$ denote the number of matched bits between $\kappa_m$  and $\hat{\kappa}_n$. Under the null hypothesis $H_0$ (no valid watermark present), the number of matched bits $S$ follows a Binomial distribution, i.e., $S \sim \text{Binomial}(M, \frac{1}{2})$ because bit matches are expected to be random in this case. For a target frame-level false positive rate $\gamma_f$, we compute:
\begin{equation}
\tau_f = \min\left\{\tau : \Pr(S \geq \tau \mid H_0) \leq \gamma_f\right\}.
\end{equation}
\noindent A frame-level assignment is considered valid only if $S_{\pi_i, \rho_i} \geq \tau_f$. Let $\mathcal{Q} = \{(\pi_i, \rho_i)|S_{\pi_i, \rho_i} \geq \tau_f\}_{i=1}^{|\mathcal{Q}|} \subseteq \mathcal{M}$ be the set of valid assignments, where $|\mathcal{Q}|$ is the number of valid assignments. Let $p_f = \Pr(S \geq \tau_f \mid H_0)$. Under $H_0$, the number of valid assignments also follows a Binomial distribution, i.e., $|\mathcal{Q}| \sim \text{Binomial}(|\mathcal{M}|, p_f)$. For a target video-level false positive rate $\gamma_v$, we compute:   
\begin{equation}
\tau_v = \min\left\{\tau : \Pr(|\mathcal{Q}| \geq \tau \mid H_0) \leq \gamma_v\right\}.
\end{equation}
\noindent The given video $\mathbf{x}^{*}$  is deemed to contain a valid watermark only if $|\mathcal{Q}| \geq \tau_v$. Finally, it must be noted that the set of valid assignments $\mathcal{Q}$ provides the information necessary to localize temporal modifications in a watermarked video.

%% file: sec/4_experiments.tex
\begin{table*}[t]
\centering
    \caption{Video quality and watermark detection performance of \texttt{SPDMark} 
    compared to existing video watermarking methods. All values are averaged across test videos. For \texttt{SPDMark}, payload is reported as bits per frame $\times$ number of frames.
    $\uparrow$ indicates higher is better. The best result in each column is shown in \textbf{bold} and the second is \underline{underlined}.}
    \begin{tabular}{llcccccc}
    \toprule
    \multirow{2.5}{*}{\textbf{Model}} & 
    \multirow{2.5}{*}{\textbf{Method}} &
    \multicolumn{2}{c}{\textbf{Extraction}} &
    \multicolumn{4}{c}{\textbf{Video Quality}} \\
    \cmidrule(lr){3-4} \cmidrule(lr){5-8}
     & & \textbf{Payload} & \textbf{Bit Acc $\uparrow$} 
  & \textbf{SC $\uparrow$} 
     & \textbf{BC $\uparrow$} & \textbf{MS $\uparrow$} & \textbf{IQ $\uparrow$} \\
    \midrule
    \multirow{4.5}{*}{\textbf{SVD-XT}} 
     & VideoShield & $512$ & $0.979$  & $0.954$ & \underline{$0.954$} & $0.956$ & $\mathbf{0.695}$ \\
     & VideoSeal & $256$ & $\mathbf{0.999}$ & \underline{$0.955$} & $0.950$ & \underline{$0.961$} & $0.682$ \\
     & VidSig & $48$ & $0.958$ &  $0.951$ & $0.953$ & $0.956$ & \underline{$0.693$} \\
     & \bgb\textbf{\texttt{SPDMark} (Ours)} & \bgb $28\times25$ & \underline{\bgb $0.995$}  & \bgb  $\mathbf{0.966}$ & \bgb $\mathbf{ 0.958}$ & \bgb $\mathbf{0.975}$ & \bgb $0.690$ \\

    \midrule

    \multirow{4.5}{*}{\textbf{ModelScope (MS)}} 
     & VideoShield & $512$ & $\mathbf{1.000}$ &  $0.927$ & $0.954$ & $0.963$ & $0.612$ \\
     & VideoSeal & $256$ & $0.999$ & \underline{$0.944$} & \underline{$0.965$} & \underline{$0.964$} & \underline{$0.614$} \\
     & VidSig & $48$ & $\mathbf{1.000}$  & $0.932$ & $0.955$ & $0.963$ & $0.610$ \\
          & \bgb \textbf{\texttt{SPDMark} (Ours)} & \bgb $28\times16$ & \bgb $0.988$ & \bgb $\mathbf{ 0.948}$ & \bgb $\mathbf{ 0.968}$ & \bgb $\mathbf{ 0.972}$ & \bgb $\mathbf{ 0.623}$ \\ 
    \bottomrule
    \end{tabular}
    \label{tab:quality}
    \vspace{-1em}
\end{table*}

\section{Experiments}
\subsection{Implementation Details}
\paragraph{Base Models.}
We implement \texttt{SPDMark} on two video diffusion architectures:
\textit{Stable Video Diffusion (SVD-XT)}~\cite{Stablevideodiffusion}: an image-to-video model configured for $576{\times}1024$ resolution with $25$ frames at $7$\,fps and $25$ inference steps; 
\textit{ModelScope (MS)}~\cite{wang2023modelscope}: a text-to-video model operating at $256{\times}256$ resolution with 16 frames and $50$ denoising steps.
For MS, we evaluate on $50$ prompts from the VBench test split~\cite{huang2023vbench} covering five categories (Animal, Human, Plant, Scenery, Vehicles; 10 each), generating $4$ videos per prompt ($50{\times}4{=}200$ videos).
For SVD-XT, we first synthesize $200$ conditioning images with a T2I model (LDM v1.5\cite{rombach2022high}) using the same $50$ prompts, then feed these images to SVD-XT to obtain a matched set of $200$ i2v videos. We train on 10{,}000 videos~from OpenVid-1M~\cite{openvid}, additional training details are provided in the Appendix.

\noindent \textbf{Basis-selection configuration.}
We attach learned basis shifts to the latent decoder’s $L{=}14$ spatial ResNet blocks.
Each block has $P{=}4$ parallel low-rank adapters (rank $r{=}32$), giving $\log_2 P{=}2$ bits per block and a payload of 28 bits per frame.

\noindent \textbf{Extractor $\mathcal{V}_{\eta}$.} We use a frame-wise ResNet-50~\cite{deepResidual} (ImageNet-pretrained \cite{deng2009imagenet}) with the final FC replaced by a linear head outputting $28$ logits per frame.
Given $v\in\mathbb{R}^{B\times T\times 3\times H\times W}$, frames are normalized to ImageNet statistics and processed independently to yield $[B,T,28]$ logits.
Training minimizes \eqref{eq:lrec} against $\kappa_t$.
At inference, we use test-time batch normalization, computing BN statistics over all $T$ frames of each test video to stabilize predictions under the length/resolution mismatch. 
Additional details on computational cost and efficiency are provided in Appendix.

\subsection{Evaluation Metrics}
\label{sec:metrics}
We evaluate \texttt{SPDMark} along three axes: imperceptibility, message recoverability, and temporal and spatial robustness.

\noindent \textbf{Generation Quality.} 
We report VBench~\cite{huang2023vbench} metrics: Subject Consistency (SC), Background Consistency (BC), Motion Smoothness (MS), and Imaging Quality (IQ) using the released evaluator. Additional details on these metrics are provided in the Appendix.

\noindent \textbf{Watermark Extraction.}
Given the valid aligned pairs $\mathcal{Q}=\{(\pi_i,\rho_i)\}$, we compute
\begin{equation}
    \begin{aligned}
        \text{Bit Acc} &= \frac{1}{|\mathcal{Q}|}\sum_{(\pi,\rho)\in\mathcal{Q}}\frac{S_{\pi,\rho}}{M},  \\
        \text{Order Acc} &= \frac{1}{|\mathcal{Q}|-1}\sum_{i}\mathbb{I}[\rho_i<\rho_{i+1}]. 
    \end{aligned}
\end{equation}

For temporal attacks, we additionally report precision, recall, and F1 over modified frame indices.

\subsection{Robustness and Attack Protocol}

To evaluate robustness under realistic video degradation, we test \texttt{SPDMark} across three categories of attacks: (i) photometric and spatial distortions (Gaussian noise, blur, geometric transforms, compression artifacts), (ii) temporal manipulations (frame drops, swaps, inserts, trims), and (iii) real-world post-processing such as video recompression, screen-recording-style degradation, and STTN-based inpainting regeneration~\citep{STTN}. These attacks capture common transformations involved in sharing, editing, or platform transcoding.  Full attack definitions, parameter settings, and implementation details are provided in the Appendix.

\begin{figure*}[t]
\centering
\setlength{\tabcolsep}{2pt}
\footnotesize

\begin{tabular}{c *{5}{cc}}
& \multicolumn{2}{c}{\textbf{No Watermark}}
& \multicolumn{2}{c}{\textbf{\texttt{SPDMark}(Ours)}}
& \multicolumn{2}{c}{\textbf{VideoShield}}
& \multicolumn{2}{c}{\textbf{VideoSeal}}
& \multicolumn{2}{c}{\textbf{VidSig}} \\
\addlinespace[2pt]

& \includegraphics[width=0.085\linewidth]{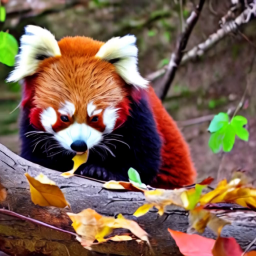}
& \includegraphics[width=0.085\linewidth]{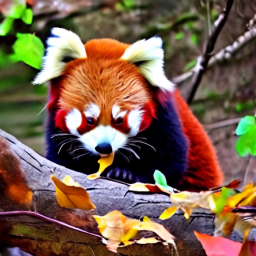}
& \includegraphics[width=0.085\linewidth]{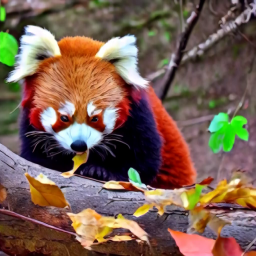}
& \includegraphics[width=0.085\linewidth]{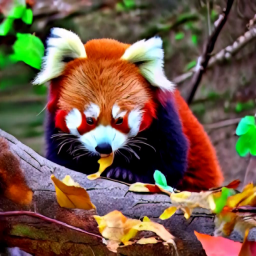}
& \includegraphics[width=0.085\linewidth]{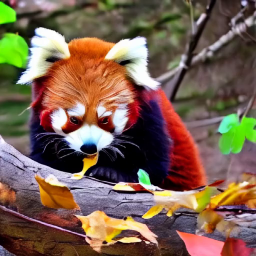}
& \includegraphics[width=0.085\linewidth]{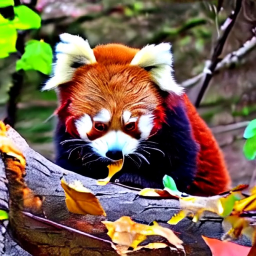}
& \includegraphics[width=0.085\linewidth]{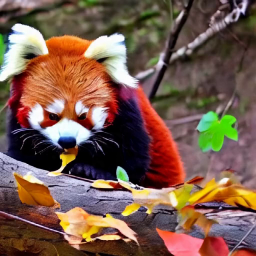}
& \includegraphics[width=0.085\linewidth]{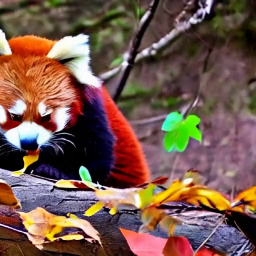}
& \includegraphics[width=0.085\linewidth]{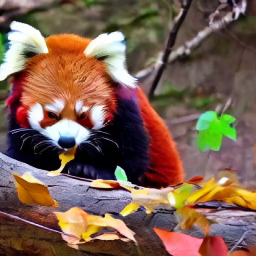}
& \includegraphics[width=0.085\linewidth]{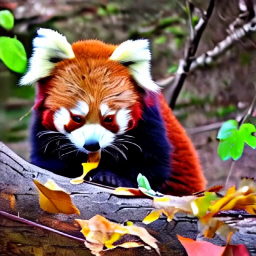}
\\[6pt]

& \includegraphics[width=0.085\linewidth]{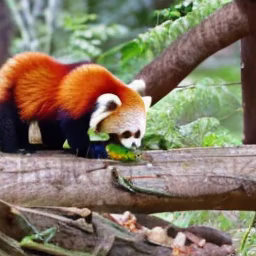}
& \includegraphics[width=0.085\linewidth]{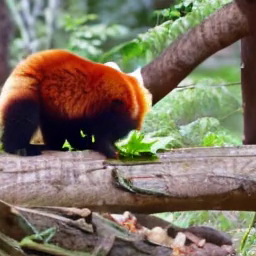}
& \includegraphics[width=0.085\linewidth]{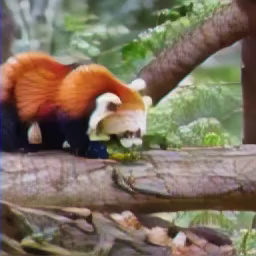}
& \includegraphics[width=0.085\linewidth]{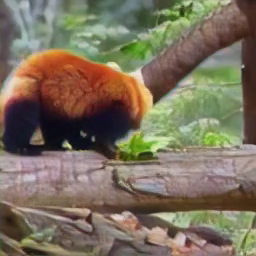}
& \includegraphics[width=0.085\linewidth]{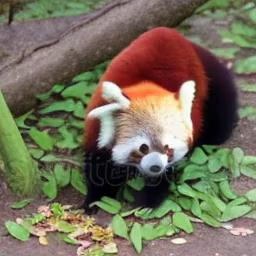}
& \includegraphics[width=0.085\linewidth]{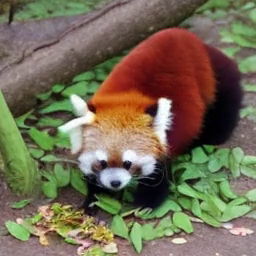}
& \includegraphics[width=0.085\linewidth]{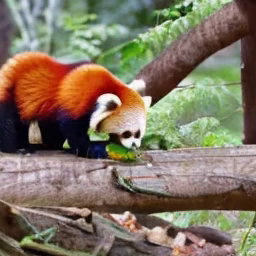}
& \includegraphics[width=0.085\linewidth]{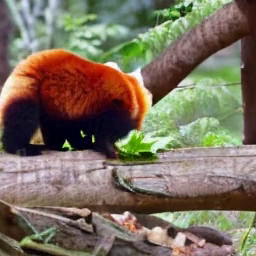}
& \includegraphics[width=0.085\linewidth]{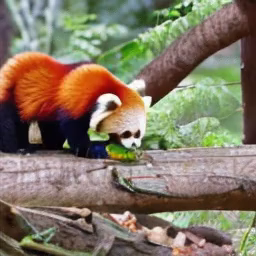}
& \includegraphics[width=0.085\linewidth]{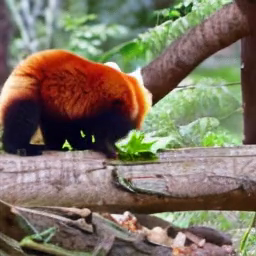}
\end{tabular}

\caption{Visual comparison across watermarking methods on SVD-XT(first row) and ModelScope(second row). For each method, we show two representative frames from the generated video.}
\label{fig:qualitative}
\end{figure*}

\begin{table*}[t]
\centering
\caption{Robustness of \textbf{\texttt{SPDMark}} under photometric, temporal, and real-world post-processing attacks. Metrics report average watermark Bit Acc ($\uparrow$). The best average robust result is shown in \textbf{bold} and the second best is \underline{underlined}.}
\resizebox{\textwidth}{!}{
\begin{tabular}{l|ccccccccc}
\toprule 
\rowcolor{lightred}\multicolumn{10}{l}{\textit{SVD-XT} \hspace{21em} (I: Photometric and Spatial Attacks, II: Temporal Tampering, III: Post-Processing)} \\ \midrule 
\textbf{Method $(\downarrow) / $ Attack $(\rightarrow)$} & I. Gauss & I. Blur & I. Crop & I. Rotate & I. Rescale & I. ColorJit & II. Drop & II. Swap & \\ \midrule 
VideoShield & $0.964$ & $0.962$ & $0.521$ & $0.507$ & $0.976$ & $0.978$ & -- & $0.935$ & -- \\ 
VideoSeal & $0.972$ & $0.998$ & $0.996$ & $0.687$ & $0.999$ & $0.999$ & $0.999$ & $0.999$ & -- \\ 
VidSig & $0.937$ & $0.420$ & $0.438$ & $0.521$ & $0.688$ & $0.958$ & $0.958$ & $0.958$ & -- \\ 
\rowcolor{lightred}\textbf{\texttt{SPDMark} (Ours)} & $0.958$ & $0.847$ & $0.989$ & $0.930$ & $0.910$ & $0.993$ & $0.988$ & $0.994$ & -- \\ \midrule 
\textbf{Method/Attack} & II. Insert & II. Trim & III. Recomp & III. ScreenRec & III. Denoise & III. STTN & III. Subtitle & III. Crop\&Drop & Avg. \\ \midrule 
VideoShield & --      & --      & $0.906$ & $0.653$ & $0.961$ & $0.991$ & $0.973$ & $0.501$ & $0.833$ \\ 
VideoSeal & $0.999$ & $0.998$ & $0.838$ & $0.598$ & $0.999$ & $0.999$ & $0.999$ & $0.513$ & \underline{$0.912$} \\ 
VidSig & $0.958$ & $0.937$ & $0.521$ & $0.563$ & $0.438$ & $0.708$ & $0.500$ & $0.458$ & $0.685$ \\ 
\rowcolor{lightred}\textbf{\texttt{SPDMark} (Ours)} & $0.994$ & $0.993$ & $0.880$ & $0.837$ & $0.921$ & $0.878$ & $0.993$ & $0.856$ & $\mathbf{0.935}$ \\ \midrule 
\midrule 
\rowcolor{lightblue}\multicolumn{10}{l}{\textit{ModelScope (MS)} \hspace{17em} (I: Photometric and Spatial Attacks, II: Temporal Tampering, III: Post-Processing)} \\ \midrule 
\textbf{Method/Attack} & I. Gauss & I. Blur & I. Crop & I. Rotate & I. Rescale & I. ColorJit & II. Drop & II. Swap & \\ \midrule 
VideoShield & $1.000$ & $0.999$ & $0.651$ & $0.523$ & $1.000$ & $1.000$ & --      & $0.997$ & -- \\ 
VideoSeal & $0.913$ & $0.951$ & $0.998$ & $0.972$ & $0.976$ & $0.999$ & $0.990$ & $0.999$ & -- \\ 
VidSig & $0.937$ & $0.542$ & $0.958$ & $0.771$ & $0.563$ & $0.979$ & $1.000$ & $1.000$ & -- \\ 
\rowcolor{lightblue}\textbf{\texttt{SPDMark} (Ours)} & $0.953$ & $0.839$ & $0.978$ & $0.941$ & $0.882$ & $0.981$ & $0.969$ & $0.986$ & -- \\ \midrule 
\textbf{Method/Attack} & II. Insert & II. Trim & III. Recomp & III. ScreenRec & III. Denoise & III. STTN & III. Subtitle & III. Crop\&Drop & Avg. \\ \midrule 
VideoShield             & --      & --      & $0.998$ & $0.853$ & $1.000$ & $1.000$ & $1.000$ & $0.500$ & $0.886$ \\ 
VideoSeal               & $0.999$ & $0.998$ & $0.531$ & $0.511$ & $0.999$ & $0.999$ & $0.982$ & $0.674$ & \underline{$0.906$} \\ 
VidSig                  & $1.000$ & $1.000$ & $0.521$ & $0.400$ & $0.688$ & $0.979$ & $1.000$ & $0.646$ & $0.812$ \\ 
\rowcolor{lightblue}\textbf{\texttt{SPDMark} (Ours)} & $0.986$ & $0.984$ & $0.986$ & $0.821$ & $0.902$ & $0.976$ & $0.981$ & $0.866$ & $\mathbf{0.939}$ \\ \bottomrule 
\end{tabular}
\vspace{-1em}
}

\label{tab:robustness_full}
\end{table*}

\subsection{Results}
\noindent \textbf{Watermark Extraction and Generation Quality.} 
We compare \texttt{SPDMark} against VideoShield, VideoSeal, and VidSig. Table~\ref{tab:quality} reports extraction accuracy and video quality metrics.
On SVD-XT, \texttt{SPDMark} achieves a bit accuracy of $99.5\%$, comparable to the strongest baselines, and maintains stable performance on ModelScope ($98.8\%$). Across video quality metrics, \texttt{SPDMark} performs competitively: it achieves the highest subject consistency (SC) and motion smoothness (MS) on both SVD-XT and ModelScope, indicating that the learned routing patterns preserve the underlying temporal structure. On ModelScope, \texttt{SPDMark} also attains the strongest background consistency (BC) and image quality (IQ), while on SVD-XT its BC and IQ remain competitive with the best baselines. Overall, \texttt{SPDMark} preserves semantic content and scene layout (Fig.~\ref{fig:qualitative}). Additional qualitative examples and extended comparisons are provided in the Appendix. 

\noindent \textbf{Watermark Robustness.} Table~\ref{tab:robustness_full} summarizes robustness across 16 photometric, geometric, temporal, and post-processing attacks. On average, \texttt{SPDMark} achieves $93.5\%$ robustness on SVD-XT and $93.9\%$ on ModelScope, outperforming VideoShield, VidSig, and VideoSeal overall. Across many photometric and temporal distortions, \texttt{SPDMark} maintains high extraction accuracy, with particularly strong performance under cropping, color jitter, and frame-level manipulations. Robustness is lower for stronger degradation-based attacks such as blur, rescaling, and denoising on ModelScope, but \texttt{SPDMark} still retains strong average performance. Under geometric transformations such as rotation and cropping, \texttt{SPDMark} substantially outperforms noise-space approaches like VideoShield, indicating improved spatial stability. For frame-level manipulations (drops, swaps, insertions, and trims), \texttt{SPDMark} achieves high extraction rates, aided by the per-frame alignment verification. Under recompression, denoising, and screen recording, \texttt{SPDMark} remains competitive with or stronger than other in-generation methods, while VideoSeal occasionally performs better under certain compression settings. Overall, the results indicate that \texttt{SPDMark} remains robust across a diverse range of perturbations, with particular strength in geometric and temporal settings.

\begin{table}[t]
\centering
\caption{Temporal forensics performance of \texttt{SPDMark} under temporal attacks. Frame Drop/Insert are evaluated with Precision, Recall, and F1 ($\uparrow$), 
while Swap attacks use Order Accuracy ($\uparrow$).} 
\label{tab:temporal_forensics}
\resizebox{\columnwidth}{!}{
\begin{tabular}{lcc}
\toprule
\rowcolor{lightgray}\textbf{Attack} &
\multicolumn{1}{c}{\textbf{SVD-XT}} &
\multicolumn{1}{c}{\textbf{ModelScope (MS)}} \\ 
\midrule
\rowcolor{lightblue}\multicolumn{3}{l}{\textit{Frame-altering Attacks (Prec./Rec./F1)}} \\
Frame Drop   & $0.999$ / $0.999$ / $0.999$ & $0.998$ / $0.999$ / $0.999$ \\
Frame Insert & $0.996$ / $0.994$ / $0.995$ & $0.998$ / $0.999$ / $0.999$ \\
\midrule
\rowcolor{lightblue}\multicolumn{3}{l}{\textit{Order-altering Attacks (Order Accuracy)}} \\ 
Random Swap   & \multicolumn{1}{c}{$1.000$} & \multicolumn{1}{c}{$1.000$} \\
Adjacent Swap & \multicolumn{1}{c}{$1.000$} & \multicolumn{1}{c}{$1.000$} \\
\bottomrule
\end{tabular}
}
\end{table}

\subsection{Ablation Studies}
\label{sec:ablation}

\begin{figure}[t]
  \centering
  \includegraphics[width=\columnwidth]{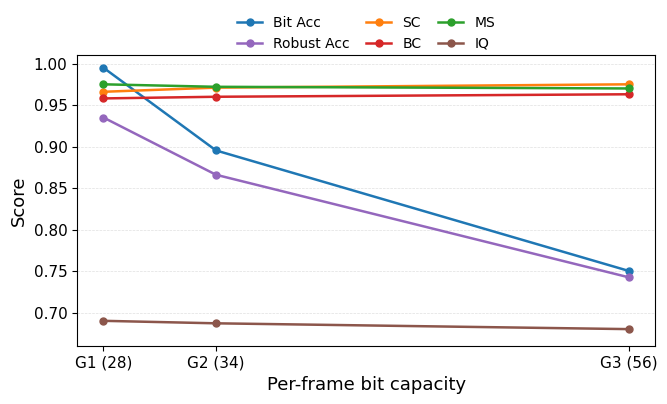}
  \caption{Bit Acc, Robust Acc, SC, BC, MS, and IQ vs.\ per-frame bit length for G1-G3, all trained with identical compute (training steps, hyperparameters and hardware).}
  \label{fig:metrics-vs-bits}
\end{figure}
\subsubsection{Basis-Selection Placement and Bit Capacity}
\label{sec:layerSelection}
We study how the placement and number of basis shift sites affect per-frame capacity and fidelity.
In our decoder, $L$ spatial ResNet blocks and one attention block (Q/K/V) are potential sites. We compare three placements (all with $P{=}4$): \textbf{G1} uses one site per ResNet block (shared conv1/conv2), $L{=}14$ $\Rightarrow$ $M{=}28$ bits/frame; \textbf{G2} adds the three attention projections, $L{=}17$ $\Rightarrow$ $M{=}34$; \textbf{G3} treats conv1 and conv2 as separate sites, $L{=}28$ $\Rightarrow$ $M{=}56$. Figure~\ref{fig:metrics-vs-bits} shows results for the three configurations. Under a fixed training budget, increasing capacity from 28 bits (G1) to 56 bits (G3) reduces bit and robustness accuracies, whereas quality metrics show minimal variation. 
To assess whether the observed payload trade-off arises solely from increased capacity or is also influenced by optimization budget, we continue training the higher-payload variants for additional steps. Specifically, we extend training by $+6$k steps for G2 and $+12$k steps for G3, improving clean/robust accuracy to $0.919/0.880$ for G2 and $0.969/0.890$ for G3. These results suggest that the degradation observed under the fixed-budget setting is partly an optimization effect, and that higher-payload configurations can recover a substantial portion of extraction robustness when given additional training.

\noindent \textbf{Sampling configurations.}
Table~\ref{tab:sampling-rank} reports an ablation on SVD-XT covering the number of generated frames, diffusion steps, and classifier-free guidance scale. 
Across all settings, \texttt{SPDMark} maintains high extraction reliability, with bit accuracy remaining above 94\% for short clips (8 frames) and reaching 99\% or higher for 16–32 frames. Robust accuracy shows a modest improvement with longer videos (from $0.880$ at 8 frames to $0.933$ at 32 frames), which is expected since additional frames provide more opportunities for correct alignment in the verification stage. Varying the number of diffusion steps (10 vs.\ 50) and guidance strength (1 vs.\ 5) has a limited influence on watermark recovery or video-quality metrics, indicating that the basis-shift perturbations applied in the decoder interact weakly with the sampling procedure. Additional results for ModelScope are provided in the Appendix.
\begin{table}[t]
\centering
\small
\caption{Sampling and ablation studies on \textbf{SVD-XT}. Robust Acc.\ denotes average bit accuracy under the attack suite used in Table~\ref{tab:robustness_full}.}
\label{tab:sampling-rank}
\setlength{\tabcolsep}{4pt}
\renewcommand{\arraystretch}{1.15}
\resizebox{\columnwidth}{!}{

\begin{tabular}{l c c c c c c c}
\toprule
\rowcolor{lightgray}\textbf{Factor} & \textbf{Setting}
& \textbf{Bit Acc ↑} & \textbf{Robust Acc ↑}
& \textbf{SC ↑} & \textbf{BC ↑}
& \textbf{MS ↑} & \textbf{IQ ↑} \\
\midrule

\multirow{3}{*}{Frames}
& $8$  & $0.942$ & $0.880$ & \underline{$0.976$} & $\mathbf{0.968}$ & \underline{$0.970$} & $\mathbf{0.721}$ \\
& $16$ & \underline{$0.989$} & \underline{$0.913$} & $\mathbf{0.978}$ & $0.962$ & $\mathbf{0.972}$ & \underline{$0.710$} \\
& $32$ & $\mathbf{0.996}$ & $\mathbf{0.933}$ & $0.965$ & $0.959$ & $0.961$ & $0.700$ \\
\midrule

\multirow{2}{*}{Steps}
& $10$ & $0.989$ & $0.902$ & $0.951$ & $0.948$ & $\mathbf{0.970}$ & $0.630$ \\
& $50$ & $\mathbf{0.995}$ & $\mathbf{0.911}$ & $\mathbf{0.971}$ & $\mathbf{0.958}$ & $0.968$ & $\mathbf{0.713}$ \\
\midrule

\multirow{2}{*}{Guidance}
& ${1}$  & $\mathbf{0.998}$ & $\mathbf{0.915}$ & $0.923$ & $0.940$ & $0.962$ & $0.640$ \\
& ${5}$ & $0.991$ & $0.910$ & $\mathbf{0.971}$ & $\mathbf{0.960}$ & $\mathbf{0.967}$ & $\mathbf{0.719}$ \\
\bottomrule
\end{tabular}}
\end{table}

%% file: sec/5_conclusion.tex
\section{Conclusion}

We presented \texttt{SPDMark}, an in-generation video watermarking framework based on Selective Parameter Displacement. By modeling watermarks as key-conditioned combination of learned low-rank basis shifts, \texttt{SPDMark} enables efficient multi-key embedding without per-key retraining. Our approach avoids the computational burden and fragility of DDIM inversion, provides per-frame key control, and integrates watermarking directly within the generative process. It enables robust extraction and forensic analysis. \texttt{SPDMark} demonstrates that carefully designed parameter displacement provides an effective and practical watermarking mechanism for modern video diffusion models, offering a compelling balance between detection accuracy, temporal coherence, and computational efficiency. Future directions could extend this work to other generative modalities.

\newpage 

%% file: sec/X_suppl.tex
\clearpage

\twocolumn[{
  \tableofcontents
}]
\newpage

\setcounter{page}{1}
\maketitlesupplementary

\appendix

\section{Datasets and training} 
We train on 10{,}000 videos~from OpenVid-1M~\cite{openvid} using AdamW 
($\beta_1{=}0.9$, $\beta_2{=}0.999$, weight decay $10^{-2}$), learning rate $10^{-4}$,
batch size 1 on 1 NVIDIA A6000 GPU, for 6000 steps. For the first 2k steps, we optimize only the message recovery loss $\mathcal{L}_{rec}$; from 2k steps onward, we optimize $L_{\mathrm{rec}} + L_{\mathrm{imp}}$.
Training uses 8-frame clips at $256$ resolution. At test time, we evaluate full-length generations (25 frames for SVD-XT and 16 frames for MS).

\section{Computational cost} 
Training takes $\approx 8$ GPU hours.
Only one basis per block is active at inference, so the decoding cost matches a single rank-$r$ low-rank update per targeted block.
The added parameters are ${\approx}2$M. Empirically, this adds ${<}5\%$ to the decoding time versus the frozen decoder.

\section{Generation Quality Evaluation Metrics}

\textbf{Subject Consistency (SC).}
For a video with frames $1,\dots,T$, let $d_t$ be the L2-normalized DINO~\citep{dino} image feature of frame $t$.
SC averages the cosine similarity of each frame to (i) the first frame and (ii) its previous frame:
\[
S_{\mathrm{SC}}
= \frac{1}{T-1} \sum_{t=2}^{T} \frac{1}{2}\big(\langle d_1, d_t\rangle + \langle d_{t-1}, d_t\rangle\big).
\]

\noindent\textbf{Background Consistency (BC).}
Let $c_t$ be the L2-normalized CLIP image feature of frame $t$. BC mirrors SC but uses CLIP features~\citep{clip}:
\[
S_{\mathrm{BC}}
= \frac{1}{T-1} \sum_{t=2}^{T} \frac{1}{2}\big(\langle c_1, c_t\rangle + \langle c_{t-1}, c_t\rangle\big).
\]

\noindent\textbf{Motion Smoothness (MS).}
Drop the odd frames to form a lower-FPS sequence and synthesize them with a video frame-interpolation model (AMT)~\citep{AMT}. For each removed frame $f_{2t-1}$ with interpolation $\hat f_{2t-1}$, compute the mean absolute error (MAE). The raw error is then normalized to $[0,1]$ (same normalization as the flicker metric):
\[
E = \frac{1}{T/2} \sum_{t=1}^{T/2} \operatorname{MAE}\!\left(\hat f_{2t-1}, f_{2t-1}\right),\qquad
S_{\mathrm{MS}} = \frac{255 - E}{255}.
\]

\noindent\textbf{Imaging Quality (IQ).}
Per frame, run the MUSIQ~\citep{ke2021musiq} image-quality predictor (0–100), then average over frames and linearly rescale:
\[
S_{\mathrm{IQ}} = \frac{1}{T}\sum_{t=1}^{T} \frac{\mathrm{MUSIQ}(t)}{100}.
\]

\section{Robustness} 
We evaluate SPDMark across photometric, temporal, and post-processing attacks.

\subsection{Attack Protocol}
\noindent \paragraph{Photometric and Spatial Attacks.}
We simulate common visual distortions encountered during content sharing: \textbf{Gaussian Noise}: additive noise $\mathcal{N}(0,\sigma^2)$ with $\sigma=0.05$. \textbf{Gaussian Blur}: Gaussian blur with an $11\times11$ kernel and $\sigma=2.0$. \textbf{Rotation}: rotation by $15^\circ$, followed by resizing/cropping back to the original dimensions. \textbf{Center Crop}: retain the central 90\% of the frame in both height and width, then resize it back to the original resolution. \textbf{Rescale}: downsample by $0.5\times$ using bicubic interpolation, then upsample back. \textbf{Color Jitter}: random brightness and contrast adjustments within $\pm 10\%$. \textbf{Subtitle}: add a semi-transparent caption box with overlaid text near the bottom of each frame, simulating subtitle or caption overlays.

\noindent \paragraph{Post-Processing and Screen Recording Simulation.}
We include transformations approximating recompression pipelines and phone screen capture: \textbf{Multi-Stage Recompression}: two-pass encoding: (1) H.264 at CRF=28, then (2) decode and re-encode with H.265 at a target bitrate of 600 kbps. \textbf{Screen Recording}: approximate screen capture by downscaling to 70\% resolution and upscaling back, adding Gaussian noise ($\sigma=0.03$), applying mild vignetting, and finally recompressing at 600 kbps. \textbf{Denoising}: apply mild Gaussian smoothing followed by small affine jitter, approximating denoising and stabilization-style post-processing. 
\textbf{STTN Inpainting}: We apply pretrained STTN-based video inpainting to a masked rectangular region in each video. Frames are resized to $432\times240$, the masked region is regenerated from neighboring and reference frames, and the inpainted result is composited back and resized to the original resolution.

\noindent \paragraph{Temporal Attacks.}
To evaluate temporal integrity and forensic capabilities: \textbf{Frame Drop}: randomly delete 50\% of frames uniformly. \textbf{Frame Swap (Random)}: apply random permutation through pairwise swaps. \textbf{Frame Swap (Adjacent)}: swap selected adjacent frame pairs. \textbf{Frame Insert}: insert a single frame at a random position, either by duplicating a neighboring frame or by inserting a random noise frame. \textbf{Video Trim}: remove two frames from the beginning and two frames from the end of the video.

\subsection{Frame Regeneration Attacks}
\label{sec:regen_attacks}

To further evaluate robustness against regeneration attacks, we consider frame-level regeneration scenarios in which a fraction of frames in the video is regenerated using either diffusion-based editing~\citep{regenration} or VAE-based compression pipelines~\citep{balle2018variational, cheng2020learned}. Table~\ref{tab:regen_attacks} reports bit accuracy when 30\%, 50\%, 70\%, or 100\% of the frames are regenerated. Across all settings, watermark detection remains successful, while bit accuracy degrades gradually as a larger fraction of frames is regenerated. This behavior is expected because regeneration changes the visual evidence available to the frame-wise extractor, but the video-level verification procedure can still accumulate enough valid frame matches to detect the watermark.

\begin{table}[t]
\centering
\small
\caption{Robustness under frame regeneration attacks. Values report bit accuracy when 30\%, 50\%, 70\%, or 100\% of frames are regenerated. Watermark detection remained successful in all settings.}
\label{tab:regen_attacks}
\resizebox{\columnwidth}{!}{
\begin{tabular}{lcccc}
\toprule
\rowcolor{lightgray}\textbf{Attack setting} & \textbf{30\%} & \textbf{50\%} & \textbf{70\%} & \textbf{100\%} \\
\midrule
Diffusion regeneration (step=60) & $0.961$ & $0.935$ & $0.866$ & $0.802$ \\
Diffusion regeneration (step=30) & $0.961$ & $0.925$ & $0.858$ & $0.807$ \\
Compression~\citep{balle2018variational} ($q=4$) & $0.976$ & $0.953$ & $0.891$ & $0.797$ \\
Compression~\citep{balle2018variational} ($q=8$) & $0.937$ & $0.892$ & $0.851$ & $0.874$ \\
Compression~\citep{cheng2020learned} ($q=3$) & $0.977$ & $0.950$ & $0.895$ & $0.788$ \\
Compression~\citep{cheng2020learned} ($q=6$) & $0.961$ & $0.915$ & $0.855$ & $0.829$ \\
\bottomrule
\end{tabular}}
\vspace{-0.75em}
\end{table}

\subsection{Model-Level Attacks}
\label{sec:modellevel}

We additionally evaluate \texttt{SPDMark} under model-level changes. First, we plug the watermark encoder and extractor trained on the base SVD model directly into the SVD-XT pipeline, which corresponds to a denoiser-level change since SVD-XT is a fine-tuned variant of SVD. In this setting, the watermark remains recoverable with Bit Acc.~$=0.987$ and Robust Acc.~$=0.909$, while preserving generation quality (SC/BC/MS/IQ $= 0.964/0.958/0.963/0.680$).  To further test robustness across denoiser architectures, we apply the same trained watermark encoder and extractor to Latte, which uses a Transformer-based denoiser. Even without retraining, the watermark remains recoverable with Bit Acc.~$=0.9790$ and Robust Acc.~$=0.9079$, with SC/BC/MS/IQ $= 0.984/0.979/0.973/0.619$.  We also evaluate post-training quantization of the VAE decoder by simulating INT8 per-channel weight quantization using round-and-dequantize. Under this quantization, \texttt{SPDMark} remains recoverable with Bit Acc.~$=0.9797$ and Robust Acc.~$=0.9035$, indicating resilience to moderate deployment-time quantization.

In contrast, when the adversary fine-tunes the VAE decoder while keeping the learned basis shifts frozen, watermark extraction fails (Bit Acc.~$\approx 0.65$). This behavior is expected because \texttt{SPDMark} relies on the decoder weights remaining consistent with the learned basis-shift dictionary. Overall, these results suggest that \texttt{SPDMark} is robust to denoiser-side changes and post-training quantization in provider-controlled deployments, but not to adversarial retraining of the decoder itself.

\subsection{Sensitivity to Detection Thresholds}
\label{app:threshold_sensitivity}

SPDMark uses frame-level and video-level verification thresholds, denoted by $(\gamma_f,\gamma_v)$, during alignment-based watermark detection. To assess sensitivity to these parameters, we vary both thresholds over a small grid of operating points and report three quantities: \emph{No Watermark}, which measures the false positive rate on non-watermarked videos; \emph{Watermark}, which measures the true positive rate on clean watermarked videos; and \emph{Robust}, which measures the average true positive rate on attacked watermarked videos.

\begin{figure}[t]
    \centering
    \includegraphics[width=\linewidth]{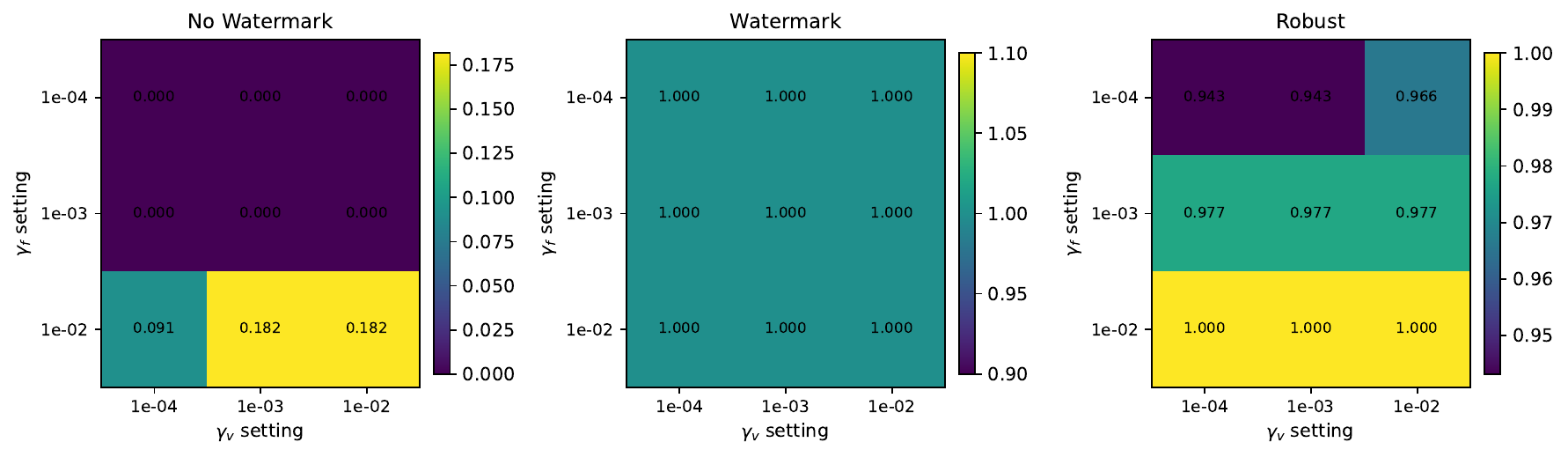}
    \caption{Detection behavior of SPDMark under different threshold settings $(\gamma_f,\gamma_v)$. No Watermark reports the false positive rate on non-watermarked videos, Watermark reports the true positive rate on clean watermarked videos, and Robust reports the average true positive rate on attacked watermarked videos.}
    \label{fig:threshold_sensitivity}
\end{figure}

Figure~\ref{fig:threshold_sensitivity} shows that SPDMark is stable across a broad range of threshold settings. In particular, the \emph{Watermark} true positive rate remains at 100\% for all tested values of $(\gamma_f,\gamma_v)$. The \emph{Robust} true positive rate also remains high, ranging from 94.3\% to 100.0\%. As expected, looser thresholds slightly increase the \emph{No Watermark} false positive rate; however, it remains at 0\% for the stricter and default settings.

\section{Additional Qualitative Examples}

\begin{figure*}[t]
\centering
\setlength{\tabcolsep}{1pt}
\footnotesize

\newcommand{\vidimgw}{0.13\linewidth} 

\begin{tabular}{*{7}{c}}
\multicolumn{7}{c}{\textbf{SVD}}\\[2pt]

\includegraphics[width=\vidimgw]{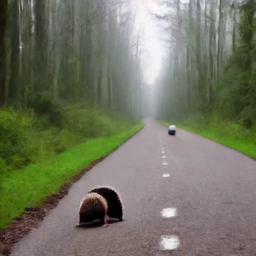} &
\includegraphics[width=\vidimgw]{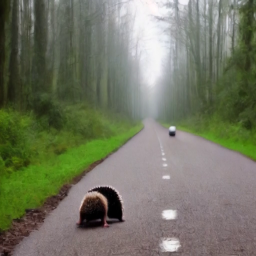} &
\includegraphics[width=\vidimgw]{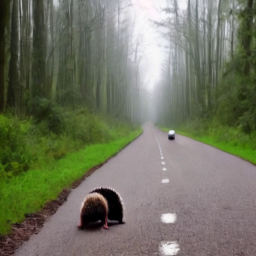} &
\includegraphics[width=\vidimgw]{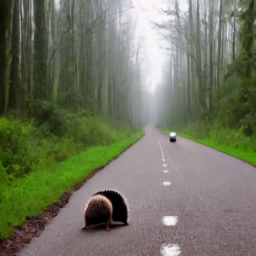} &
\includegraphics[width=\vidimgw]{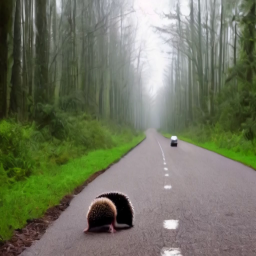} &
\includegraphics[width=\vidimgw]{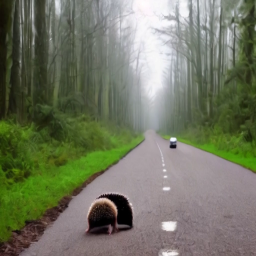} &
\includegraphics[width=\vidimgw]{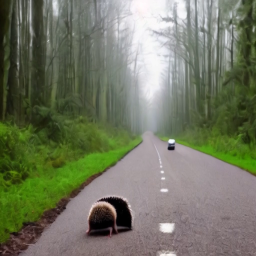} \\[-2pt]

\includegraphics[width=\vidimgw]{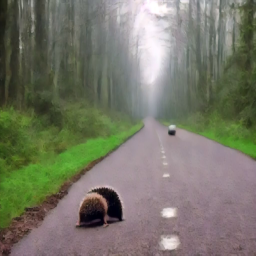} &
\includegraphics[width=\vidimgw]{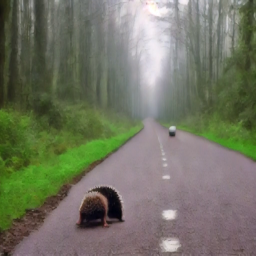} &
\includegraphics[width=\vidimgw]{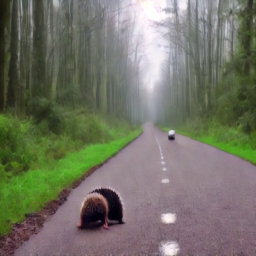} &
\includegraphics[width=\vidimgw]{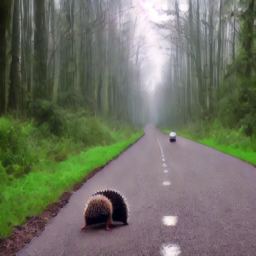} &
\includegraphics[width=\vidimgw]{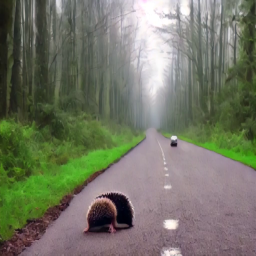} &
\includegraphics[width=\vidimgw]{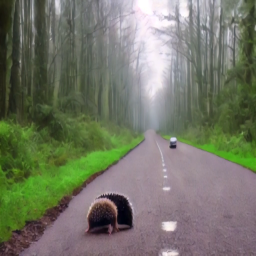} &
\includegraphics[width=\vidimgw]{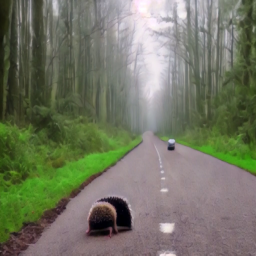} \\[4pt]

\includegraphics[width=\vidimgw]{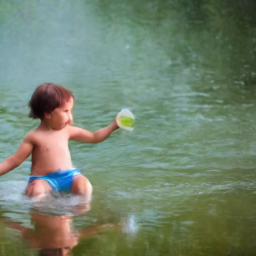} &
\includegraphics[width=\vidimgw]{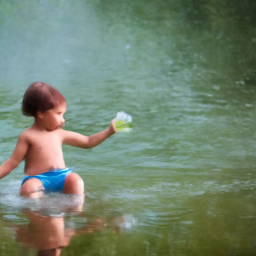} &
\includegraphics[width=\vidimgw]{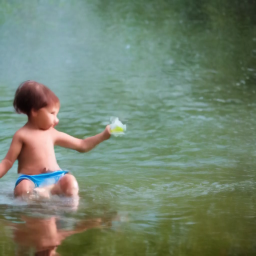} &
\includegraphics[width=\vidimgw]{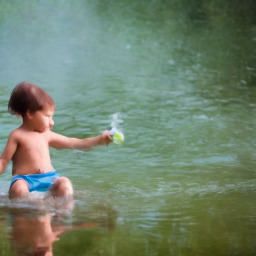} &
\includegraphics[width=\vidimgw]{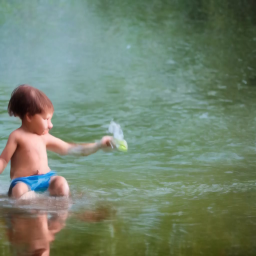} &
\includegraphics[width=\vidimgw]{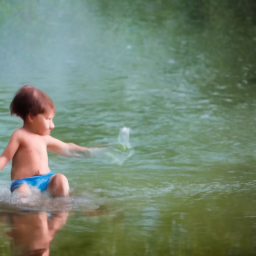} &
\includegraphics[width=\vidimgw]{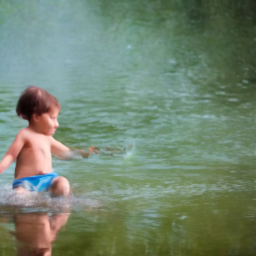} \\[-2pt]

\includegraphics[width=\vidimgw]{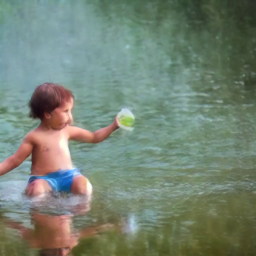} &
\includegraphics[width=\vidimgw]{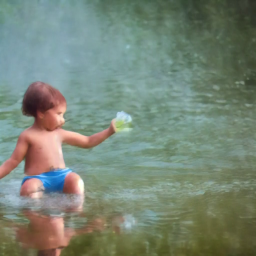} &
\includegraphics[width=\vidimgw]{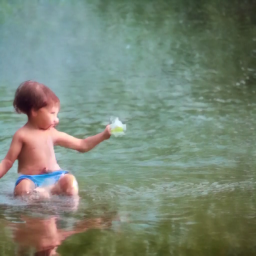} &
\includegraphics[width=\vidimgw]{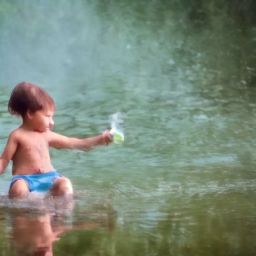} &
\includegraphics[width=\vidimgw]{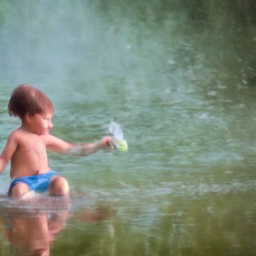} &
\includegraphics[width=\vidimgw]{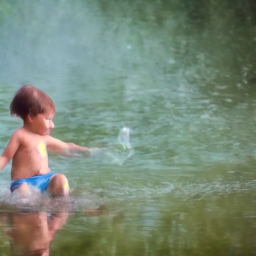} &
\includegraphics[width=\vidimgw]{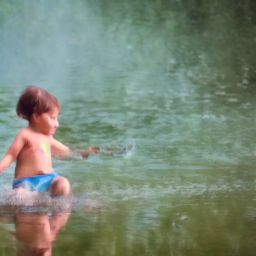} \\[4pt]

\includegraphics[width=\vidimgw]{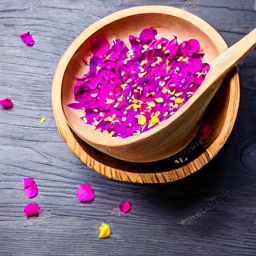} &
\includegraphics[width=\vidimgw]{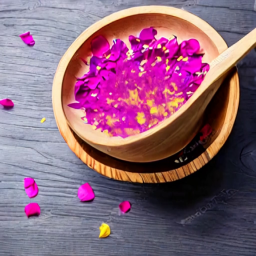} &
\includegraphics[width=\vidimgw]{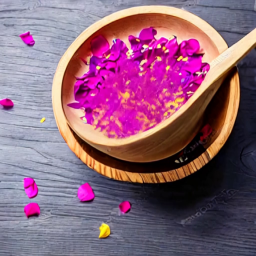} &
\includegraphics[width=\vidimgw]{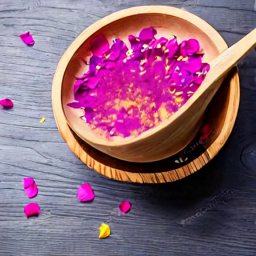} &
\includegraphics[width=\vidimgw]{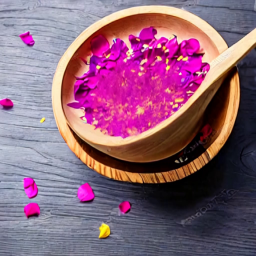} &
\includegraphics[width=\vidimgw]{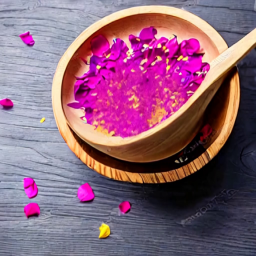} &
\includegraphics[width=\vidimgw]{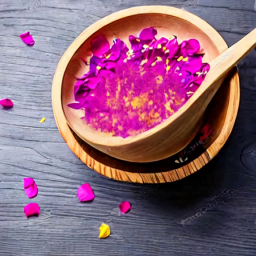} \\[-2pt]

\includegraphics[width=\vidimgw]{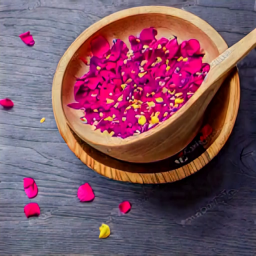} &
\includegraphics[width=\vidimgw]{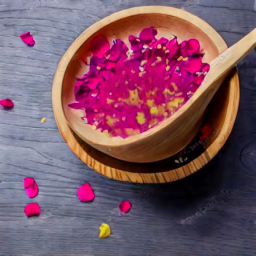} &
\includegraphics[width=\vidimgw]{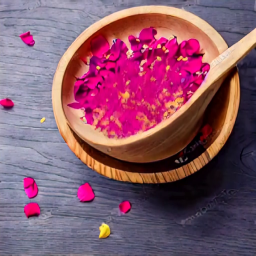} &
\includegraphics[width=\vidimgw]{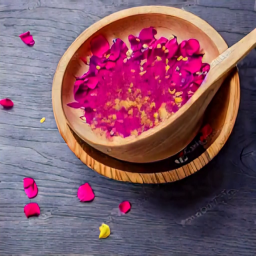} &
\includegraphics[width=\vidimgw]{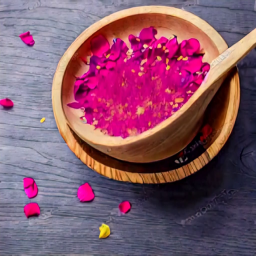} &
\includegraphics[width=\vidimgw]{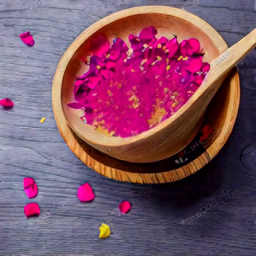} &
\includegraphics[width=\vidimgw]{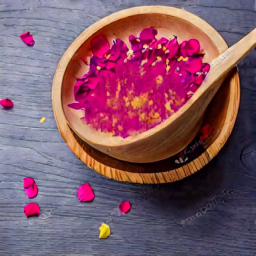} \\

\end{tabular}

\caption{SVD videos: Seven frames per row. For each video, the No watermark row is followed by the corresponding SPDMark row (for the same video).}
\label{fig:svd_alternating_rows}
\vspace{-0.5em}
\end{figure*}

\begin{figure*}[t]
\centering
\setlength{\tabcolsep}{1pt}
\footnotesize

\newcommand{\vidimgw}{0.13\linewidth} 

\begin{tabular}{*{7}{c}}
\multicolumn{7}{c}{\textbf{ModelScope}}\\[2pt]

\includegraphics[width=\vidimgw]{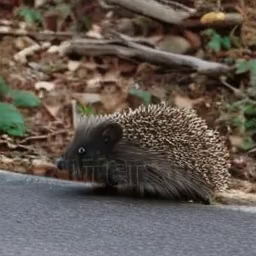} &
\includegraphics[width=\vidimgw]{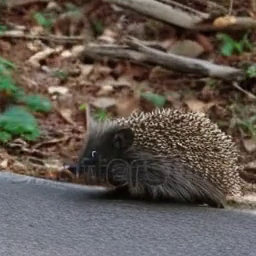} &
\includegraphics[width=\vidimgw]{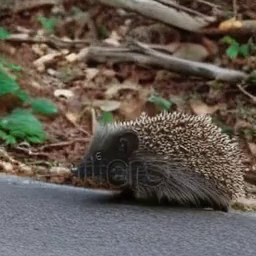} &
\includegraphics[width=\vidimgw]{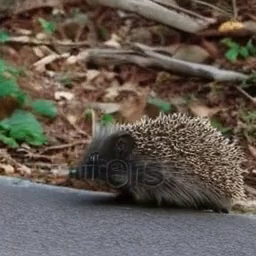} &
\includegraphics[width=\vidimgw]{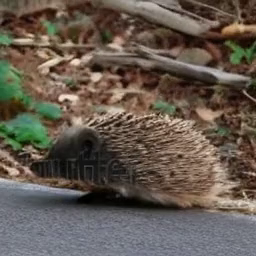} &
\includegraphics[width=\vidimgw]{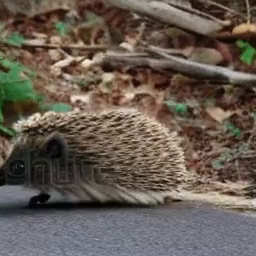} &
\includegraphics[width=\vidimgw]{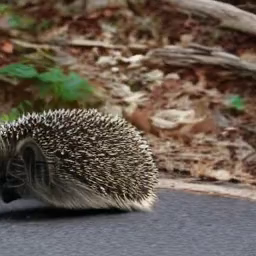} \\[-2pt]

\includegraphics[width=\vidimgw]{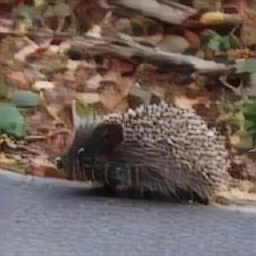} &
\includegraphics[width=\vidimgw]{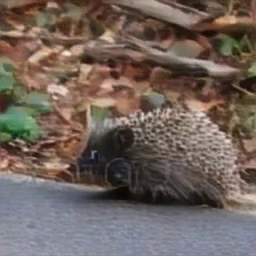} &
\includegraphics[width=\vidimgw]{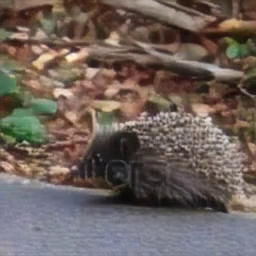} &
\includegraphics[width=\vidimgw]{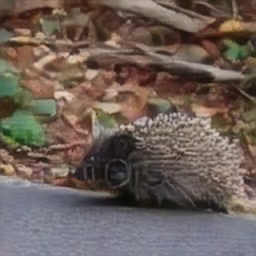} &
\includegraphics[width=\vidimgw]{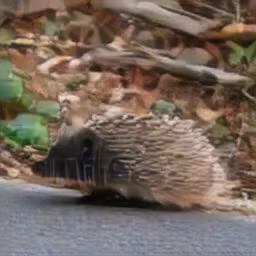} &
\includegraphics[width=\vidimgw]{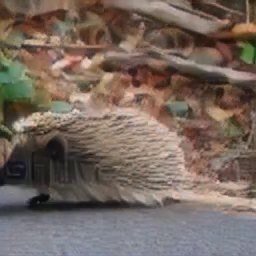} &
\includegraphics[width=\vidimgw]{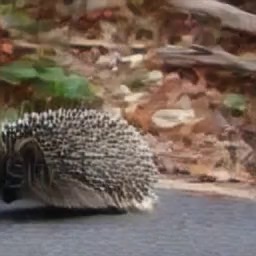} \\[4pt]

\includegraphics[width=\vidimgw]{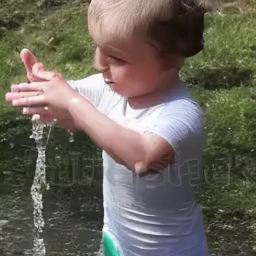} &
\includegraphics[width=\vidimgw]{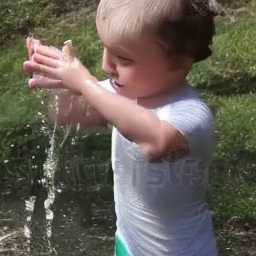} &
\includegraphics[width=\vidimgw]{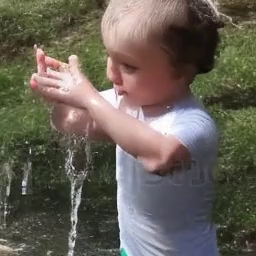} &
\includegraphics[width=\vidimgw]{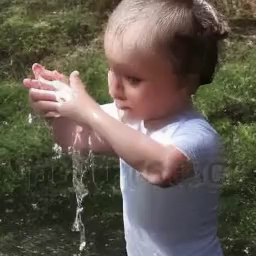} &
\includegraphics[width=\vidimgw]{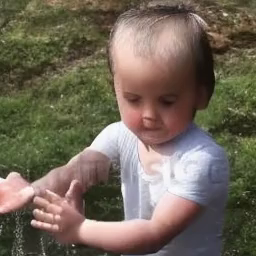} &
\includegraphics[width=\vidimgw]{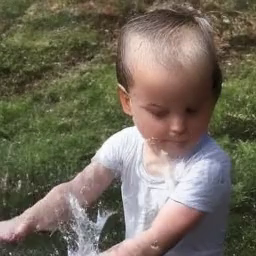} &
\includegraphics[width=\vidimgw]{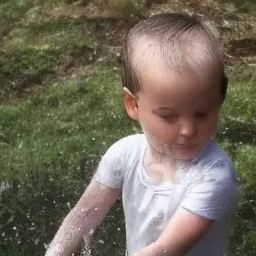} \\[-2pt]

\includegraphics[width=\vidimgw]{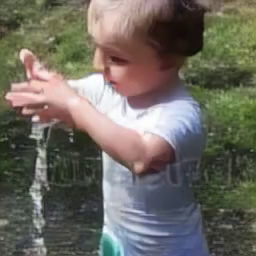} &
\includegraphics[width=\vidimgw]{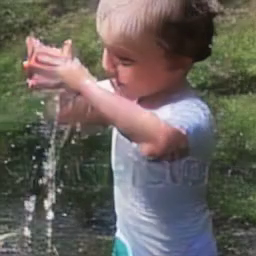} &
\includegraphics[width=\vidimgw]{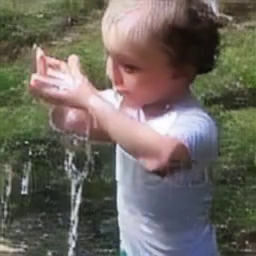} &
\includegraphics[width=\vidimgw]{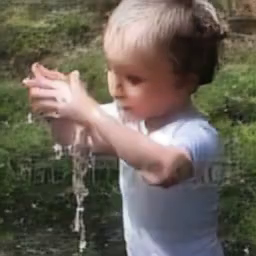} &
\includegraphics[width=\vidimgw]{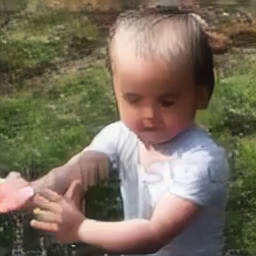} &
\includegraphics[width=\vidimgw]{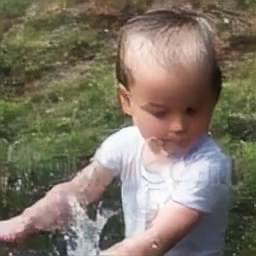} &
\includegraphics[width=\vidimgw]{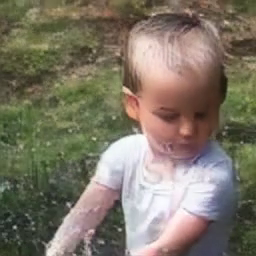} \\[4pt]

\includegraphics[width=\vidimgw]{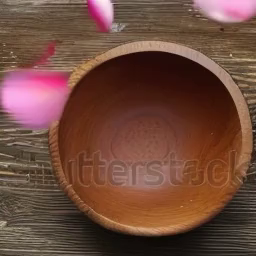} &
\includegraphics[width=\vidimgw]{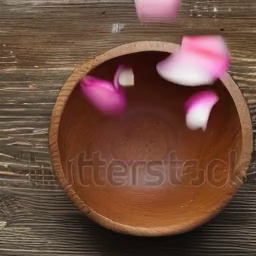} &
\includegraphics[width=\vidimgw]{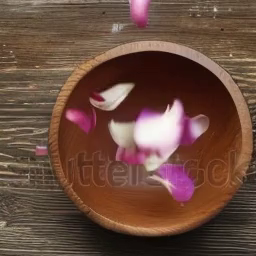} &
\includegraphics[width=\vidimgw]{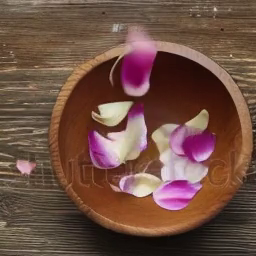} &
\includegraphics[width=\vidimgw]{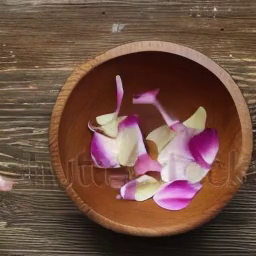} &
\includegraphics[width=\vidimgw]{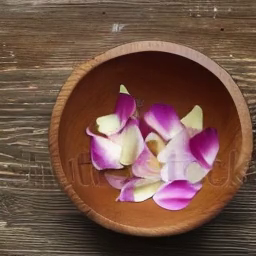} &
\includegraphics[width=\vidimgw]{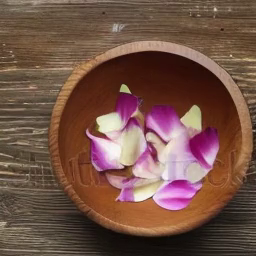} \\[-2pt]

\includegraphics[width=\vidimgw]{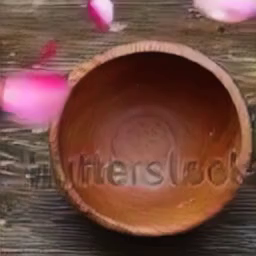} &
\includegraphics[width=\vidimgw]{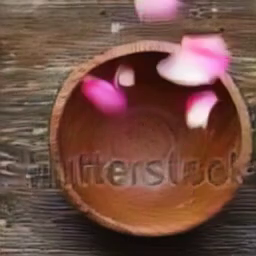} &
\includegraphics[width=\vidimgw]{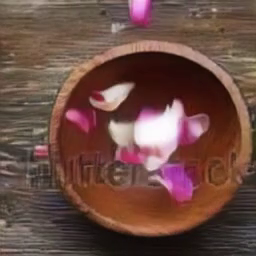} &
\includegraphics[width=\vidimgw]{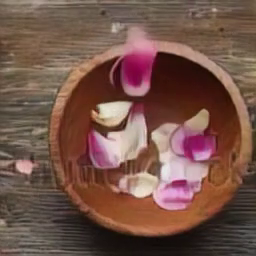} &
\includegraphics[width=\vidimgw]{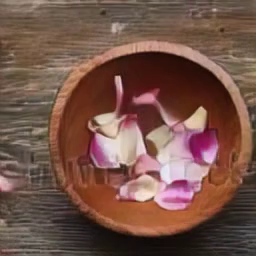} &
\includegraphics[width=\vidimgw]{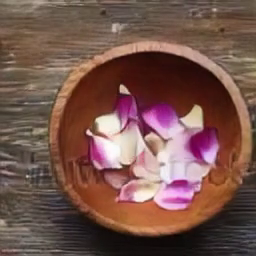} &
\includegraphics[width=\vidimgw]{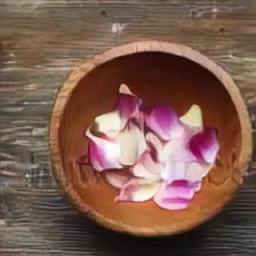} \\

\end{tabular}

\caption{ModelScope videos: Seven frames per row. For each video, the No watermark row is followed by the corresponding SPDMark row (for the same video).}
\label{fig:svd_ms_alternating_rows}
\vspace{-0.5em}
\end{figure*}

Figures~\ref{fig:svd_alternating_rows} (SVD) and \ref{fig:svd_ms_alternating_rows} (ModelScope) provide extended visualisations. For each base model, we show three videos and, for each video, display seven consecutive frames from the non-watermarked video followed by the corresponding \texttt{SPDMark} sample of the same video. texttt{SPDMark} closely tracks the clean videos: textures, edges, and colors remain visually consistent, and we do not observe watermark-induced artifacts. The sequences further indicate that temporal coherence is preserved.

\subsection{Sampling Configuration on ModelScope}
\paragraph{Setup.}
We ablate three factors for ModelScope(Table~\ref{tab:sampling-rank_scope}): number of generated frames, diffusion steps, and CFG. For each configuration, we use the same prompt set and seed protocol as in the main results, and report Bit accuracy, Robust accuracy, and Video quality metrics.
The results mirror those of SVD-XT: longer videos improve robust accuracy, while diffusion steps and guidance have a limited effect on extraction or video quality.
\begin{table}[t]
\centering
\small
\caption{Sampling and ablation studies on \textbf{ModelScope}.}
\label{tab:sampling-rank_scope}
\setlength{\tabcolsep}{4pt}
\renewcommand{\arraystretch}{1.15}
\resizebox{\columnwidth}{!}{

\begin{tabular}{l c c c c c c c}
\toprule
\rowcolor{lightgray}\textbf{Factor} & \textbf{Setting}
& \textbf{Bit Acc ↑} & \textbf{Robust Acc ↑}
& \textbf{SC ↑} & \textbf{BC ↑}
& \textbf{MS ↑} & \textbf{IQ ↑} \\
\midrule

\multirow{2}{*}{Frames}
& $8$ &  $0.880$ & $0.863$ &  $0.964$ & $0.974$& $0.974$ &  $0.480$\\
& $25$ & $0.942$ & $0.916$ &   $0.883$ & $0.936$ & $0.948$ & $0.604$\\
\midrule

\multirow{2}{*}{Steps}
& $10$ &  $0.969$ & $0.918$ &  $0.894$ &  $0.941$& $0.969$ & $0.583$\\
& $25$ & $0.977$ & $0.925$ & $0.934$ & $0.961$ & $0.972$ & $0.624$ \\
\midrule

\multirow{2}{*}{Guidance}
& ${6}$  & $0.977$ & $0.925$ &  $0.933$ & $0.963$ & $0.972$ & $0.625$ \\
& ${12}$ & $ 0.977$ &  $ 0.929$ & $0.953$ & $ 0.967$ & $0.971$ & $ 0.629$\\
\bottomrule
\end{tabular}}
\vspace{-1em}
\end{table}

%% file: main.bib
@inproceedings{VideoShield,
title={Video{S}hield: Regulating Diffusion-based Video Generation Models via Watermarking},
author={Runyi Hu and Jie Zhang and Yiming Li and Jiwei Li and Qing Guo and Han Qiu and Tianwei Zhang},
booktitle={The International Conference on Learning Representations},
year={2025},
}

@article{VideoMark,
  title={Video{M}ark: A Distortion-Free Robust Watermarking Framework for Video Diffusion Models},
  author={Hu, Xuming and Li, Hanqian and Li, Jungang and Liu, Aiwei},
  journal={arXiv preprint arXiv:2504.16359},
  year={2025}
}

@article{LVMark,
  title={{LVM}ark: Robust Watermark for Latent Video Diffusion Models},
  author={MinHyuk Jang and Youngdong Jang and JaeHyeok Lee and Feng Yang and Gyeongrok Oh and Jongheon Jeong and Sangpil Kim},
  journal={arXiv preprint arXiv:2412.09122},
  year={2025}
}

@article{VideoSignature,
  title={Video {S}ignature: In-generation Watermarking for Latent Video Diffusion Models},
  author={Huang, Yu and Chen, Junhao and Zheng, Qi and Li, Hanqian and Liu, Shuliang and Hu, Xuming},
  journal={arXiv preprint arXiv:2506.00652},
  year={2025}
}

@article{Stablevideodiffusion,
  title={Stable video diffusion: Scaling latent video diffusion models to large datasets},
  author={Blattmann, Andreas and Dockhorn, Tim and Kulal, Sumith and Mendelevitch, Daniel and Kilian, Maciej and Lorenz, Dominik and Levi, Yam and English, Zion and Voleti, Vikram and Letts, Adam and others},
  journal={arXiv preprint arXiv:2311.15127},
  year={2023}
}

@article{fernandez2023stable,
  title={The {S}table {S}ignature: Rooting Watermarks in Latent Diffusion Models},
  author={Fernandez, Pierre and Couairon, Guillaume and J{\'e}gou, Herv{\'e} and Douze, Matthijs and Furon, Teddy},
  journal={The International Conference on Computer Vision},
  year={2023}
}

@inproceedings{wen2023tree,
author = {Wen, Yuxin and Kirchenbauer, John and Geiping, Jonas and Goldstein, Tom},
title = {Tree-{R}ings watermarks: Invisible fingerprints for diffusion images},
year = {2023},
booktitle = {Advances in Neural Information Processing Systems},
}

@inproceedings{deng2009imagenet,
  title={Image{N}et: A large-scale hierarchical image database},
  author={Deng, Jia and Dong, Wei and Socher, Richard and Li, Li-Jia and Li, Kai and Fei-Fei, Li},
  booktitle={The IEEE/CVF Conference on Computer Vision and Pattern Recognition},
  year={2009},
}

@inproceedings{rombach2022high,
  title={High-resolution image synthesis with latent diffusion models},
  author={Rombach, Robin and Blattmann, Andreas and Lorenz, Dominik and Esser, Patrick and Ommer, Bj{\"o}rn},
  booktitle={The IEEE/CVF Conference on Computer Vision and Pattern Recognition},
  year={2022}
}

@article{dhariwal2021diffusion,
  title={Diffusion models beat {GANS} on image synthesis},
  author={Dhariwal, Prafulla and Nichol, Alexander},
  journal={Advances in Neural Information Processing Systems},
  year={2021}
}

@inproceedings{clip,
  title={Learning transferable visual models from natural language supervision},
  author={Radford, Alec and Kim, Jong Wook and Hallacy, Chris and Ramesh, Aditya and Goh, Gabriel and Agarwal, Sandhini and Sastry, Girish and Askell, Amanda and Mishkin, Pamela and Clark, Jack and others},
  booktitle={The International Conference on Machine Learning},
  year={2021},
}

@article{ho2020denoising,
  title={Denoising diffusion probabilistic models},
  author={Ho, Jonathan and Jain, Ajay and Abbeel, Pieter},
  journal={Advances in Neural Information Processing Systems},
  year={2020}
}

@techreport{biden2023executive,
  author       = {Biden, Joseph R.},
  title        = {Executive Order 14110 on the Safe, Secure, and Trustworthy Development and Use of {Artificial Intelligence}},
  institution  = {The White House},
  address      = {Washington, D.C.},
  year         = {2023},
  month        = {October},
  note         = {88 Fed. Reg. 75191}
}

@article{rijsbosch2025adoption,
  title={Adoption of Watermarking Measures for {AI}-Generated Content and Implications under the {EU AI} {A}ct},
  author={Rijsbosch, Bram and van Dijck, Gijs and Kollnig, Konrad},
  journal={arXiv preprint arXiv:2503.18156},
  year={2025}
}

@inproceedings{song2020denoising,
title={Denoising Diffusion Implicit Models},
author={Jiaming Song and Chenlin Meng and Stefano Ermon},
booktitle={The International Conference on Learning Representations},
year={2021},
}

@inproceedings{ronneberger2015u,
  title={U-net: Convolutional networks for biomedical image segmentation},
  author={Ronneberger, Olaf and Fischer, Philipp and Brox, Thomas},
  booktitle={The Medical Image Computing and Computer-Assisted Intervention},
  year={2015},
}

@inproceedings{deepResidual,
  title={Deep residual learning for image recognition},
  author={He, Kaiming and Zhang, Xiangyu and Ren, Shaoqing and Sun, Jian},
  booktitle={The IEEE/CVF Conference on Computer Vision and Pattern Recognition},
  year={2016}
}

@inproceedings{zhu2018hiddenhidingdatadeep,
author = {Zhu, Jiren and Kaplan, Russell and Johnson, Justin and Fei-Fei, Li},
title = {HiDDeN: Hiding Data With Deep Networks},
booktitle = {The European Conference on Computer Vision},
year = {2018},
}

@inproceedings{kim2024wouaf,
  title={{WOUAF}: Weight modulation for user attribution and fingerprinting in text-to-image diffusion models},
  author={Kim, Changhoon and Min, Kyle and Patel, Maitreya and Cheng, Sheng and Yang, Yezhou},
  booktitle={The IEEE/CVF Conference on Computer Vision and Pattern Recognition},
  year={2024}
}

@article{hu2022lora,
  title={Lo{RA}: Low-rank adaptation of large language models.},
  author={Hu, Edward J and Shen, Yelong and Wallis, Phillip and Allen-Zhu, Zeyuan and Li, Yuanzhi and Wang, Shean and Wang, Lu and Chen, Weizhu and others},
  journal={The International Conference on Learning Representations},
  year={2022}
}

@article{feng2024aqualora,
  title={Aqua{L}ora: Toward white-box protection for customized stable diffusion models via watermark {L}o{RA}},
  author={Feng, Weitao and Zhou, Wenbo and He, Jiyan and Zhang, Jie and Wei, Tianyi and Li, Guanlin and Zhang, Tianwei and Zhang, Weiming and Yu, Nenghai},
  journal={The International Conference on Machine Learning},
  year={2024}
}

@article{regenration,
  title={Invisible image watermarks are provably removable using generative {AI}},
  author={Zhao, Xuandong and Zhang, Kexun and Su, Zihao and Vasan, Saastha and Grishchenko, Ilya and Kruegel, Christopher and Vigna, Giovanni and Wang, Yu-Xiang and Li, Lei},
  journal={Advances in Neural Information Processing Systems},
  year={2024}
}

@inproceedings{yang2024gaussian,
  title={Gaussian {S}hading: Provable performance-lossless image watermarking for diffusion models},
  author={Yang, Zijin and Zeng, Kai and Chen, Kejiang and Fang, Han and Zhang, Weiming and Yu, Nenghai},
  booktitle={The IEEE/CVF Conference on Computer Vision and Pattern Recognition},
  year={2024}
}

@misc{openai2024sora,
  author = {OpenAI},
  title = {Video generation models as world simulators},
  howpublished = {\url{https://openai.com/index/video-generation-models-as-world-simulators/}},
  year = {2024},
  month = {February 15},
  note = {[Technical report]}
}

@article{wang2023modelscope,
  title={Modelscope text-to-video technical report},
  author={Wang, Jiuniu and Yuan, Hangjie and Chen, Dayou and Zhang, Yingya and Wang, Xiang and Zhang, Shiwei},
  journal={arXiv preprint arXiv:2308.06571},
  year={2023}
}

@article{bao2024vidu,
  title={Vidu: a highly consistent, dynamic and skilled text-to-video generator with diffusion models},
  author={Bao, Fan and Xiang, Chendong and Yue, Gang and He, Guande and Zhu, Hongzhou and Zheng, Kaiwen and Zhao, Min and Liu, Shilong and Wang, Yaole and Zhu, Jun},
  journal={arXiv preprint arXiv:2405.04233},
  year={2024}
}

@inproceedings{peebles2023scalableDit,
  title={Scalable diffusion models with transformers},
  author={Peebles, William and Xie, Saining},
  booktitle={The IEEE/CVF International Conference on Computer Vision},
  year={2023}
}

@article{videoSeal,
  title={Video seal: Open and efficient video watermarking},
  author={Fernandez, Pierre and Elsahar, Hady and Yalniz, I Zeki and Mourachko, Alexandre},
  journal={arXiv preprint arXiv:2412.09492},
  year={2024}
}

@article{zhang2018perceptual,
  title={The Unreasonable Effectiveness of Deep Features as a Perceptual Metric},
  author={Zhang, Richard and Isola, Phillip and Efros, Alexei A and Shechtman, Eli and Wang, Oliver},
  journal={In The IEEE/CVF Conference on Computer Vision and Pattern Recognition},
  year={2018}
}

@inproceedings{openvid,
title={Open{V}id-1{M}: A Large-Scale High-Quality Dataset for Text-to-video Generation},
author={Kepan Nan and Rui Xie and Penghao Zhou and Tiehan Fan and Zhenheng Yang and Zhijie Chen and Xiang Li and Jian Yang and Ying Tai},
booktitle={The International Conference on Learning Representations},
year={2025},
}

@InProceedings{huang2023vbench,
     title={{VBench}: Comprehensive Benchmark Suite for Video Generative Models},
     author={Huang, Ziqi and He, Yinan and Yu, Jiashuo and Zhang, Fan and Si, Chenyang and Jiang, Yuming and Zhang, Yuanhan and Wu, Tianxing and Jin, Qingyang and Chanpaisit, Nattapol and Wang, Yaohui and Chen, Xinyuan and Wang, Limin and Lin, Dahua and Qiao, Yu and Liu, Ziwei},
     booktitle={The IEEE/CVF Conference on Computer Vision and Pattern Recognition},
     year={2024}
 }

@Article{Kuhn1955,
  author  = {Kuhn, Harold W.},
  title   = {The {H}ungarian Method for the assignment problem},
  journal = {Naval Research Logistics Quarterly},
  volume  = {2},
  pages   = {83--97},
  year    = {1955},
}

@inproceedings{ke2021musiq,
  title={{MUSIQ}: Multi-scale image quality transformer},
  author={Ke, Junjie and Wang, Qifei and Wang, Yilin and Milanfar, Peyman and Yang, Feng},
  booktitle={The IEEE/CVF International Conference on Computer Vision},
  year={2021}
}

@inproceedings{dino,
  title={Emerging Properties in Self-Supervised Vision Transformers},
  author={Caron, Mathilde and Touvron, Hugo and Misra, Ishan and J\'egou, Herv\'e  and Mairal, Julien and Bojanowski, Piotr and Joulin, Armand},
  booktitle={The International Conference on Computer Vision},
  year={2021}
}

@inproceedings{AMT,
   title={{AMT}: All-Pairs Multi-Field Transforms for Efficient Frame Interpolation},
   author={Li, Zhen and Zhu, Zuo-Liang and Han, Ling-Hao and Hou, Qibin and Guo, Chun-Le and Cheng, Ming-Ming},
   booktitle={The IEEE/CVF Conference on Computer Vision and Pattern Recognition},
   year={2023}
}

@inproceedings{cheng2020learned,
  title={Learned Image Compression with Discretized {G}aussian Mixture Likelihoods and Attention Modules},
  author={Cheng, Zhengxue and Sun, Heming and Takeuchi, Masaru and Katto, Jiro},
  booktitle={The IEEE/CVF Conference on Computer Vision and Pattern Recognition},
  year={2020}
}

@inproceedings{balle2018variational,
title={Variational image compression with a scale hyperprior},
author={Johannes Ballé and David Minnen and Saurabh Singh and Sung Jin Hwang and Nick Johnston},
booktitle={The International Conference on Learning Representations},
year={2018}
}

@inproceedings{STTN,
  title={Learning joint spatial-temporal transformations for video inpainting},
  author={Zeng, Yanhong and Fu, Jianlong and Chao, Hongyang},
  booktitle={The European Conference on Computer Vision},
  year={2020},
}
